\documentclass[10pt,twocolumn,letterpaper]{article}

\usepackage{cvpr}
\usepackage{times}
\usepackage{epsfig}
\usepackage{graphicx}
\usepackage{amsmath}
\usepackage{amssymb}

\usepackage{subfigure}
\usepackage{multirow}
\usepackage{adjustbox}
\usepackage{threeparttable}

\usepackage[pagebackref=true,breaklinks=true,letterpaper=true,colorlinks,bookmarks=false]{hyperref}

\cvprfinalcopy % *** Uncomment this line for the final submission

\def\cvprPaperID{2049} % *** Enter the CVPR Paper ID here

\ifcvprfinal\fi
\begin{document}

%%%%%%%%% TITLE
\title{Dynamic Inference: A New Approach Toward Efficient Video Action Recognition}

%%%%%%%%%%%%%%%%%%%%%%%%%%%%%%%%%%%%%%%%%%%
\author{Wenhao Wu~$^{1,3}$\thanks{This work was done when Wenhao Wu was a research intern at Baidu.}\quad
Dongliang He~$^{2}$\quad
Xiao Tan~$^{2}$\quad
Shifeng Chen~$^{1}$\thanks{Corresponding author.}\quad 
Yi Yang~$^{4}$\quad
Shilei Wen~$^{2}$ \\
$^1$ Shenzhen Institutes of Advanced Technology, Chinese Academy of Sciences, China\\
$^2$ Department of Computer Vision Technology (VIS), Baidu Inc., China\\
$^3$ University of Chinese Academy of Sciences, China \\
$^4$ University of Technology Sydney, Australia
% {\tt\small \{wh.wu,shifeng.chen\}@siat.ac.cn ~~~ \{hedongliang01,tanxiao01,wenshilei\}@baidu.com } 
}
%%%%%%%%%%%%%%%%%%%%%%%%%%%%%%%%%%%%%%%%%%%

\maketitle
%\thispagestyle{empty}

%%%%%%%%% ABSTRACT
\begin{abstract}
Though action recognition in videos has achieved great success recently, it remains a challenging task due to the massive computational cost. Designing lightweight networks is a possible solution, but it may degrade the recognition performance. In this paper, we innovatively propose a general dynamic inference idea to improve inference efficiency by leveraging the variation in the distinguishability of different videos. The dynamic inference approach can be achieved from aspects of the network depth and the number of input video frames, or even in a joint input-wise and network depth-wise manner. In a nutshell, we treat input frames and network depth of the computational graph as a 2-dimensional grid, and several checkpoints are placed on this grid in advance with a prediction module. The inference is carried out progressively on the grid by following some predefined route, whenever the inference process comes across a checkpoint, an early prediction can be made depending on whether the early stop criteria meets. For the proof-of-concept purpose, we instantiate three dynamic inference frameworks using two well-known backbone CNNs. In these instances, we overcome the drawback of limited temporal coverage resulted from an early prediction by a novel frame permutation scheme, and alleviate the conflict between progressive computation and video temporal relation modeling by introducing an online temporal shift module. Extensive experiments are conducted to thoroughly analyze the effectiveness of our ideas and to inspire future research efforts. Results on various datasets also evident the superiority of our approach.

%Compared to its fixed full inference counterpart, the performance of our dynamic inference framework is equivalent while the computation during inference significantly reduced.
\end{abstract}

\begin{figure}[t]
\begin{center}
\subfigure[Different ``Writing" video instances]{
\includegraphics[width=0.9\columnwidth]{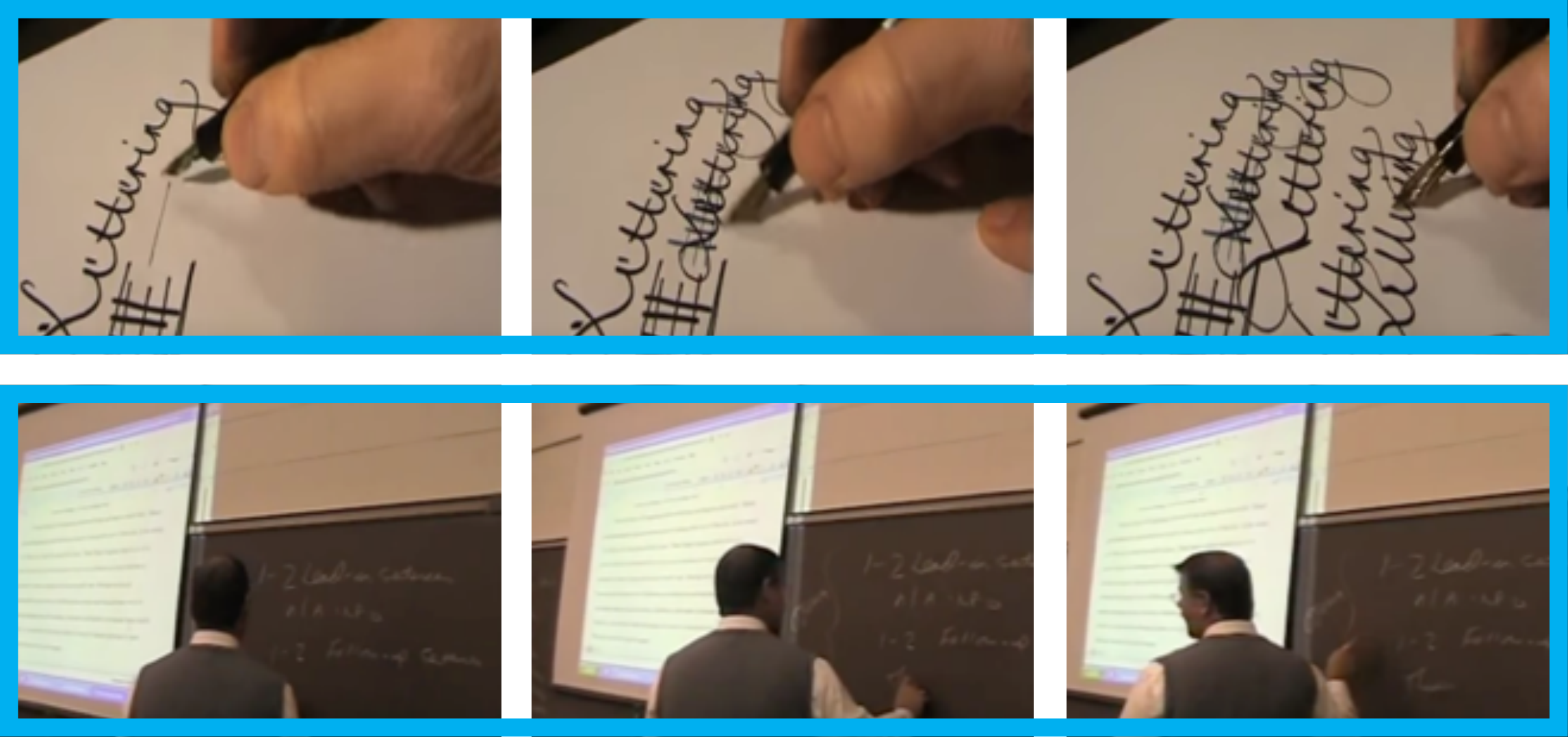}
\label{fig:depth}
}
\subfigure[``Running" v.s. ``Long Jump"]{
\includegraphics[width=0.9\columnwidth]{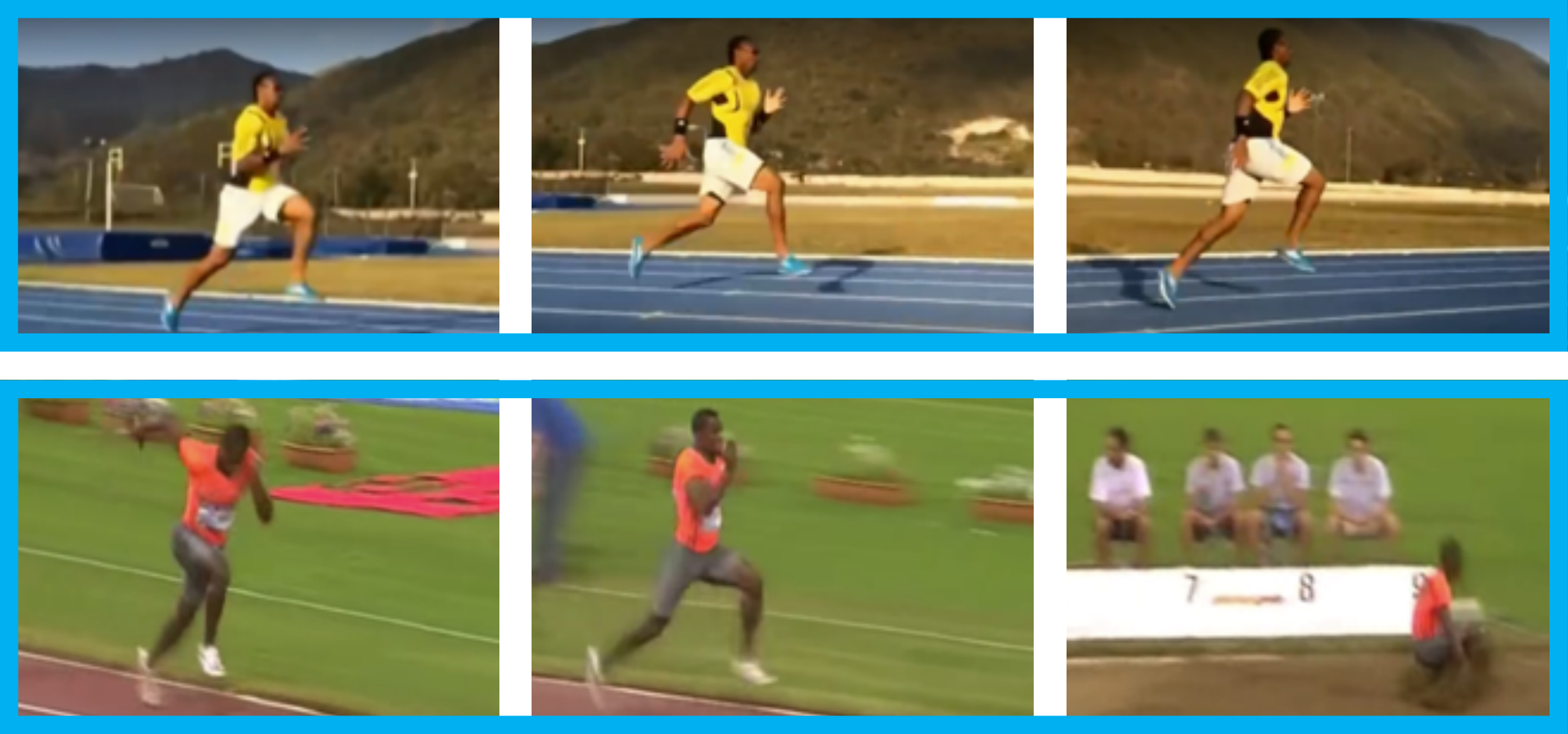}
\label{fig:input}
}
\end{center}
\caption{Illustration of diverse distinguish abilities among video instances. A few frames are sufficient to recognize ``Writing'' while watching till the end of videos is required to tell ``Running'' from ``Long Jump''. Due to the irregular viewpoint, the second video in (a) will require deeper network than the first one for feature abstraction.}
\label{fig:illustration}
%\vspace{-2 ex}
\end{figure}

%%%%%%%%% BODY TEXT
\section{Introduction}
Action recognition in videos is one of the most active research topics in the computer vision community, owing to its significant application potential for video surveillance, video recommendation, retrieval, and so on. The two most important aspects of evaluating video action recognition frameworks are classification accuracy and inference cost. Recently, significant progress has been achieved in terms of the recognition performance in this area following the deep convolutional network paradigm \cite{tsn,two-stream-stresnet,tle,T-Resnet,c3d,i3d,nonlocal,actionvlad,attentioncluster}. However, the inference efficiency remains a great challenge for large scale applications due to the heavy computation burden of deep CNN models. Therefore, in this paper, attentions are paid to the efficiency of CNN models. 

There have already existed pretty much work on improving the efficiency of action recognition in videos \cite{stnet,eco,tsm,mfnet,s3d,r2+1d,p3d,ARN,off,motion-feature}. To reduce the computation cost (\emph{aka} FLOPs, which is short for float-point operations), these works mainly focus on designing efficient network architectures to learn more representative features for classification. Meanwhile, all such solutions treat each video instance equally, i.e., all videos pass through a same CNN network route in the inference phase and a same number of frames are sampled from each video for testing, which we believe can be further improved by inference in a dynamic (or adaptive) fashion.

Our insight is that videos differentiate from each other in terms of their distinguishability, which leads to two consequences. First, varying number of frames are needed for recognizing videos. For instance, as illustrated in Fig.\ref{fig:depth}, it is quite natural for a human to tell a video belongs to ``Writing'' when a very few frames at the beginning of a video are observed. However, to tell whether a video can be categorized into ``Long Jump'' or ``Running'', we have to watch the video till its ending part (Fig.\ref{fig:input} as an example). Second, varying network capability is needed for categorizing videos due to diverse dominance of their corresponding visual features. As shown in Fig.\ref{fig:depth}, due to an irregular viewpoint, the second row is a non-typical ``Writing'' video so it demands more capable network for feature abstraction. 

Motivated by these observations, in this paper, we propose a novel idea of dynamic inference to improve action recognition efficiency. Computation resource is adaptively allocated among difference videos according to their distinguishabilities. The dynamic inference approach can be achieved from aspects of the network depth and the input video frames, or even in a joint input-wise and network depth-wise manner.
%we propose a novel framework which enables input-wise and network depth-wise dynamic inference simultaneously to allocate computation resources for video recognition adaptively. 
Specifically, we treat input frames and network blocks as a 2-dimensional grid, where we have predefined $K$ grid points as checkpoints, to each of which a prediction head is appended. 
At the inference phase, whenever the inference process comes across a checkpoint, an early prediction can be made depending on whether the early stop criteria meets. 
For the proof-of-concept purpose, dynamic inference frameworks are instantiated using the MSDNet \cite{msdnet} and ResNet \cite{resnet} backbone CNNs. In these instances, we overcome the drawback of limited temporal coverage resulted from an early prediction by a novel frame permutation scheme and alleviate the conflict between progressive computation and video temporal relation modeling by introducing an online temporal shift module \cite{tsm}. 
%In our framework, instead of sequentially feeding the input frames, we designed a novel frame permutation mechanism to enlarge the temporal coverage of input frames when it stops early and to construct diverse temporal strides for input at the same time. Besides, in order for both modeling the temporal relation of video frames efficiently and meeting the progressive computation requirement, we propose to add an online temporal shift module \cite{tsm} between temporally adjacent blocks.
Extensive experiments are conducted on multiple well-known datasets, including Kinetics-400 \cite{kay2017kinetics}, Something-Something v1 and v2 \cite{sth-sth}, UCF101 \cite{ucf101} and HMDB51 \cite{hmdb}. We empirically analyze the different behaviours of input-wise, network depth-wise as well as joint-input-depth-wise dynamic inferences and show their strength and weakness to readers for comprehensively understanding our idea. Besides, experimental results verify that our solution can significantly reduce the average FLOPs while maintaining excellent recognition accuracy, showing the superiority of dynamic inference for action recognition. Our major contributions are summarized as follows:
\begin{itemize}
    \item We are the first to improve action recognition efficiency from the dynamic inference viewpoint, which is previously neglected. Our work makes the reader think of efficient action recognition differently from lightweight model designing.
    \item For proof of concept, we turn dynamic inference idea into practical network instances by proposing frame permutation and online temporal shift to tackle the raised obstacles of limited temporal coverage and conflict between progressive computation and temporal relation modeling.
    \item Extensive empirical analysis based on the three instances shows the strength and weakness of input-wise, network depth-wise and joint dynamic inferences under different circumstances. Our models also achieve significant efficiency improvement compared to their fixed inference counterparts.  
\end{itemize}

%------------------------------------------------------------------------
\section{Related Work}
Action recognition has drawn great research attention in the community \cite{wang2017untrimmednets,wu2019multi,two-stream,tsn,T-Resnet,c3d,i3d,s3d,p3d,nonlocal,r2+1d,ARN,eco,mfnet,stnet,tsm}. Our work focuses on efficient action recognition and it is closely related to the following two lines of research jobs.

\textbf{More efficient network architectures} specifically designed for video recognition have been well studied in the literature. Following the I3D \cite{i3d} paradigm for spatial-temporal modeling, S3D \cite{s3d}, P3D \cite{p3d}, R(2+1)D \cite{r2+1d}, MFNet \cite{mfnet} and StNet \cite{stnet} are proposed to reduce computation overhead of 3D convolution while remaining the spatial-temporal modeling property. These works choose to decompose 3D convolution into 2D spatial convolution followed by 1D temporal convolution on either per convolution operation basis or per 3D convolution network block basis. There exist several other networks which merge 2D and 3D information in CNN blocks to enhance the feature abstraction capability and resort to shallower backbones for efficiencies, such as ECO \cite{eco} and ARTNet \cite{ARN}. Another research direction is to superimpose motion information learning into appearance feature network to reduce the overhead of motion stream for two-stream solution \cite{two-stream}, the design of the networks of such solutions is inspired by the calculation process of dense optical flow, typical works include optical flow guided feature network \cite{off} and motion feature network proposed in \cite{motion-feature}. All of these existing solutions can be regarded as orthogonal research directions to our dynamic inference idea, and they can be further improved with appropriate dynamic inference adaptation.

\begin{figure*}
\begin{center}
\subfigure[Depth-wise]{
\includegraphics[width=0.257\linewidth]{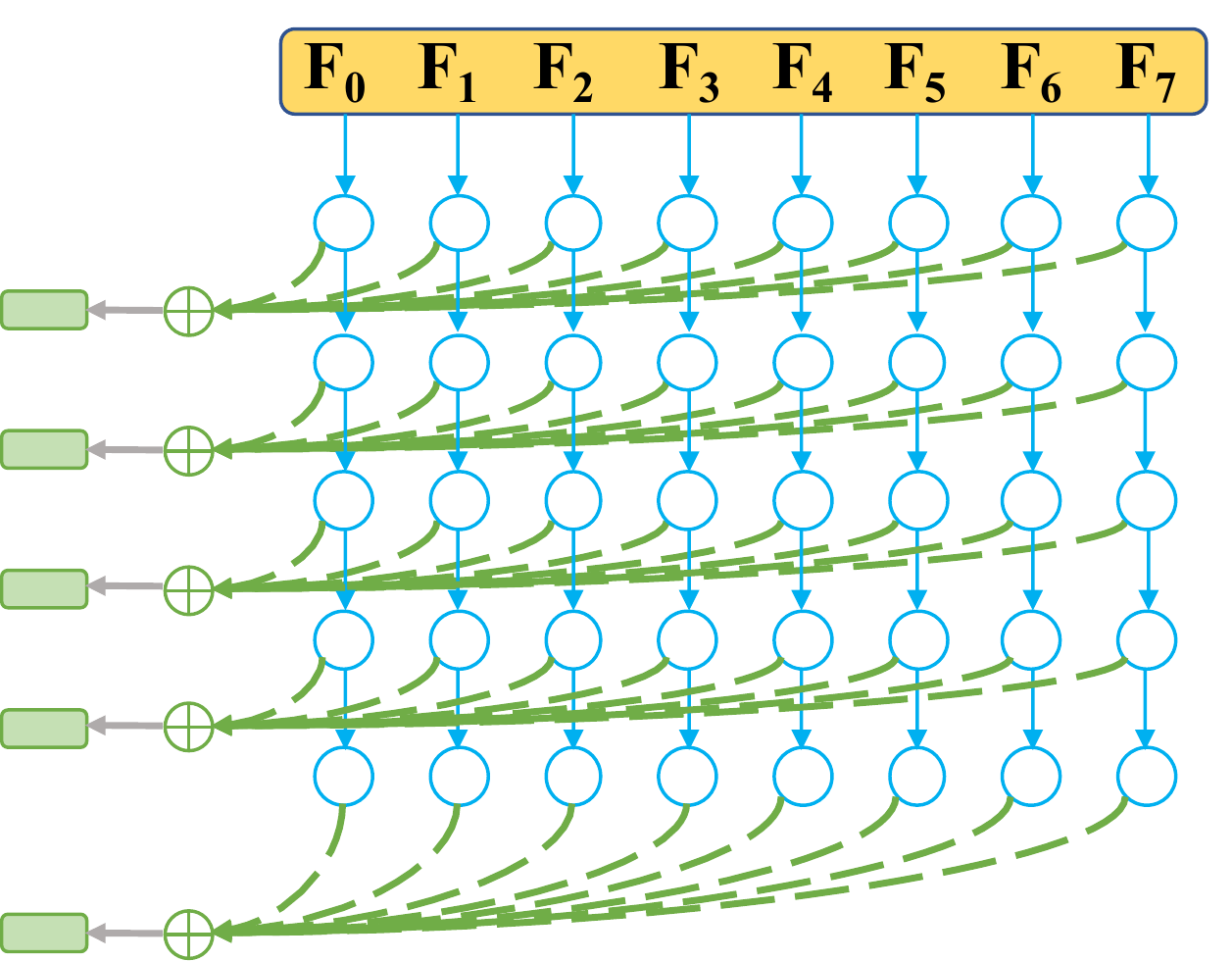}
\label{fig:addepth}
}
\subfigure[Input-wise]{
\includegraphics[width=0.205\linewidth]{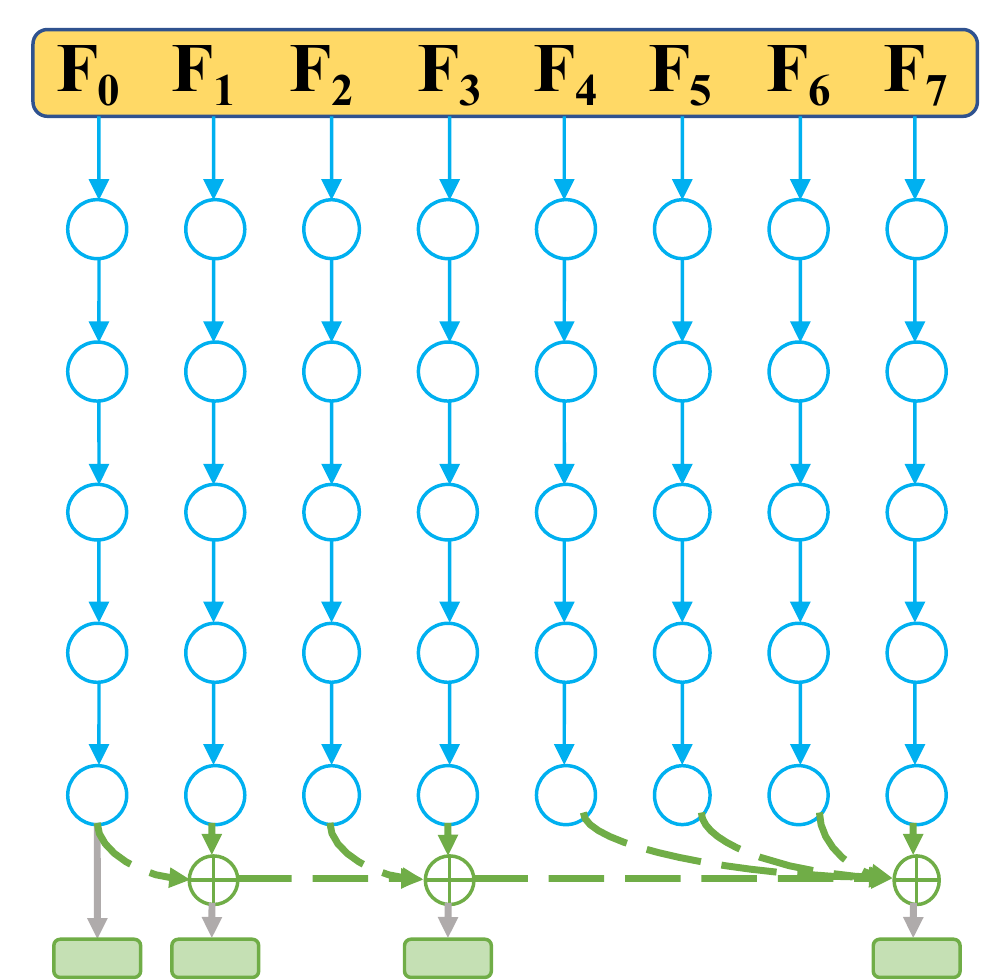}
\label{fig:adinput}
}
\subfigure[Input with permutation]{
\includegraphics[width=0.197\linewidth]{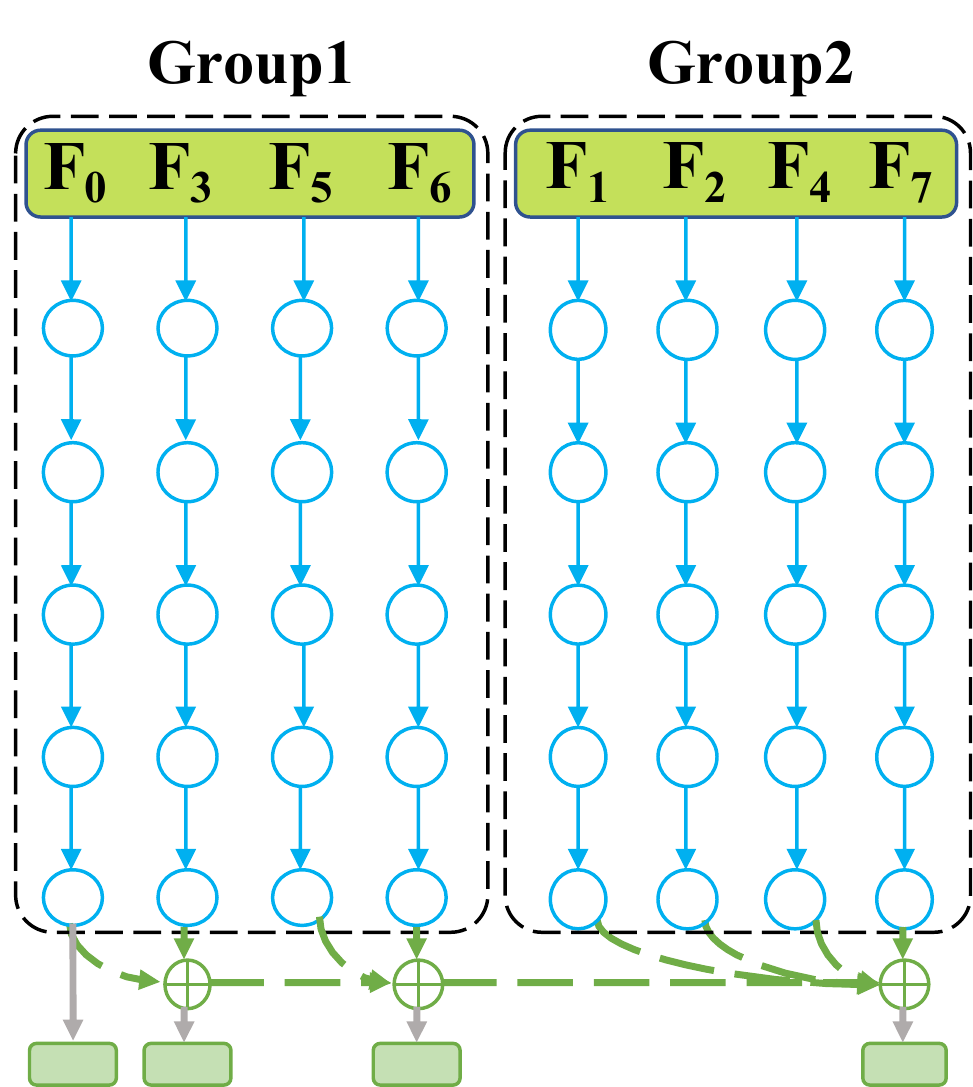}
\label{fig:admt-input}
}
\subfigure[Unified]{
\includegraphics[width=0.277\linewidth]{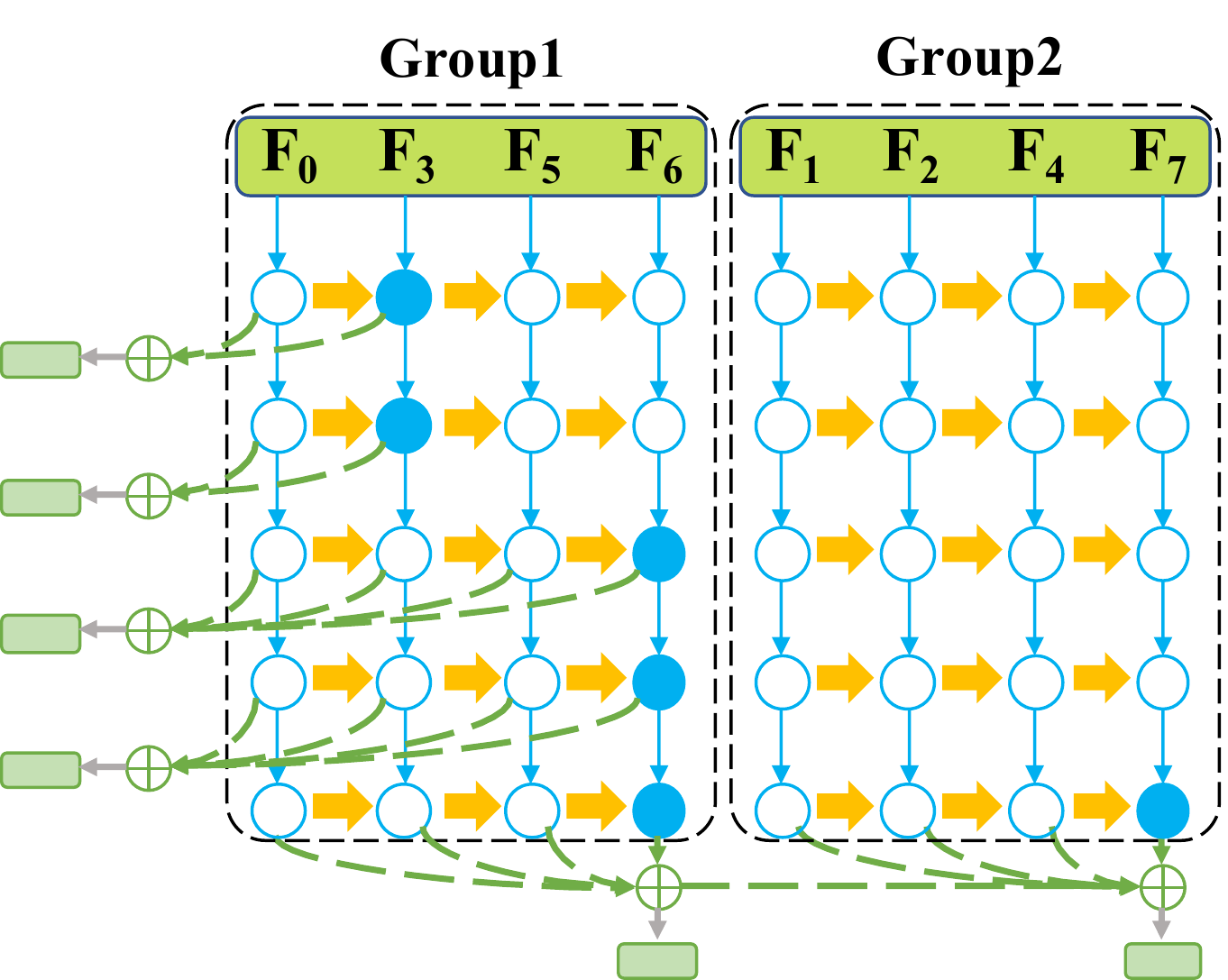}
\label{fig:unify}
}
\end{center}
\caption{Illustration of different dynamic inference schemes when five blocks and eight frames are used for recognizing a video. Blue arrows and circles denote 2D-CNN blocks and their corresponding feature maps. Green bricks mean prediction heads and yellow arrows mean online temporal shift modules. The solid blue circles represent predefined checkpoints in our framework. (a) and (b) show a naive 5-level depth-wise and 4-level input-wise dynamic inference solution, respectively. (c) depicts that with frame shuffle, each group of input frames constructs an input sequence with multiple temporal strides and the temporal coverage of the valid input is enlarged if early stop is made. (d) shows our proposed frame permutation + online temporal shift solution.} 
\label{fig:dynamic inference}
\end{figure*}

\textbf{Dynamic network route} has already been studied for the image classification task in the literature as well. In \cite{guo2016depth}, depth dropout is used in the training phase for efficient deep residual network training. The idea of stochastic depth network \cite{huang2016deep} is also leveraged as sort of regularization of deep networks in the training phase. In addition to the training phase, a dynamic network is also proposed for testing. In \cite{msdnet}, the authors propose to adaptively determine the network depth for different images and a multi-scale dense network is designed for image classification. Region level network depth adaptation is also studied for image classification in \cite{figurnov2017spatially}, in this work, computation resource is allocated among different regions of a testing image instead of among testing images. Network depth adaptation is a very effective solution for image classification, and significant inference efficiency improvement is achieved. Our work is largely inspired by these solutions for image recognition. However, as best as we can know, we are the first to leverage dynamic inference for video recognition so far. 

%------------------------------------------------------------------------
\section{Approach}
\label{sec:approach}
\subsection{Dynamic Inference Formulation}
%Videos contain multiple frames and it is a common practice to extract $l$ frames (namely, snippets) from a video for recognition, as is done in multiple state-of-the-art frameworks \cite{tsm,s3d,r2+1d,stnet}.
Dynamic inference for video recognition poses two orientations, namely input-wise and network depth-wise. Generally, action recognition can be formulated as $z=\mathcal{P}\circ\mathcal{E}(\mathcal{I}_0,\mathcal{I}_1,...,\mathcal{I}_{K-1})$, where $\mathcal{P}$ and $\mathcal{E}$ denote predictor and feature extractor, $\mathcal{I}_i$ denote the $i^{th}$ set of input frames. Under the dynamic inference scenario, the progressive prediction can, in general, be formulated as:
\begin{equation}
    \label{formulation}
    z_i = 
    \begin{cases}
    \mathcal{P}_0(y_0)=\mathcal{P}_0\circ\mathcal{E}_0(\mathcal{I}_0), & if ~i = 0 \\
    \mathcal{P}_i(y_i)=\mathcal{P}_i\circ\mathcal{E}_i(y_{i-1},y_{i-2},...,y_{0}, \mathcal{I}_i) &i>0 
    \end{cases},
\end{equation}
where $i=0,1,...,K-1$ is the index of progressive processing step (namely, checkpoint) and $y_i$ are features extracted at the $i^{th}$ step. The final prediction result $z=z_i$ when the stop criteria $C_i$ is met where $C_i$ can be defined to be a function of the previous predictions as $C_i: \phi(z_0,z_1,...,z_i) > 0$. A dynamic inference framework varies in the design of $\mathcal{E}_i$, $\mathcal{P}_i$, and the early stop criteria $C_i$.

The general dynamic inference idea can be directly specialized to input-wise or network depth-wise approaches if Eq.\ref{e1} or Eq.\ref{e2} is met, respectively.
\begin{equation}
    \label{e1}
    \begin{cases}
    y_i = \mathcal{E}_i(y_{i-1},y_{i-2},...,y_{0}, \mathcal{I}_i) = \mathcal{A}_i(\mathcal{E}_i(\mathcal{I}_i), y_{i-1},...,y_0)\\
    \mathcal{I}_i \neq None, \forall~i \geq 0
    \end{cases}
\end{equation}
\begin{equation}
    \label{e2}
    \begin{cases}
    y_i = \mathcal{E}_i(y_{i-1},y_{i-2},...,y_{0}, \mathcal{I}_i) = \mathcal{E}_i(y_{i-1},...,y_0)\\
    \mathcal{I}_i = None, ~i>0 \\
    \mathcal{I}_0 = \{F_0,F_1,...,F_{l-1}\}
    \end{cases},
\end{equation}
where $\mathcal{A}_i$ is feature aggregation function (e.g., average pooling) and $F_0,...,F_{l-1}$ are all the sampled frames and $l$ is the number of sampled frames.

From Eq.\ref{formulation}, the following concerns on designing general dynamic inference networks raise naturally:
1) when early stop at $z_i$, the sampled input frames sets, $\mathcal{I}_0,...,\mathcal{I}_i$, cover very limited part of the input video.
2) dynamic inference network should be enabled with progressive computation capability. The previous computation outputs should be utilized to support current computation incrementally. 
3) temporal relation modeling is supposed to enhance recognition performance, meanwhile temporal modeling such as 3D convolution involves temporal dependencies and it conflicts with progressive computation requirement.

\subsection{Instantiation}
Following Eq.\ref{formulation}, we can design a variety of dynamic inference frameworks. For the proof-of-concept purpose, we showcase how to instantiate dynamic frameworks by keeping the above three concerns in mind.
We treat input frames sets(denoted as $\mathcal{I}_0, \mathcal{I}_1, ..., \mathcal{I}_{N-1}$) and network blocks (denoted as $B_0, B_1, ... , B_{M-1}$) as a 2-dimensional grid, where we have predefined $K$ grid points (denoted as $\mathcal{I}_{i_0}B_{j_0}, \mathcal{I}_{i_1}B_{j_1}, ..., \mathcal{I}_{i_{K-1}}B_{j_{K-1}}, ~0\le i_0 \le i_1 \le ... \le i_{K-1} < N ~\&~ 0\le j_0 \le j_1 \le ... \le j_{K-1} < M$) as checkpoints, to each of which a prediction head is appended. 
At the inference phase, whenever a checkpoint is reached, an early prediction can be made depending on whether the stop criteria meets.
Here, the criteria for the $k^{th}$ checkpoint is defined as the hypothesis of $C_k: \max \{s_k\} > T_k$, where $s_k$ is the classification score at the $k^{th}$ checkpoint and $T_k$ is a threshold. 
Fig.\ref{fig:addepth} and Fig.\ref{fig:adinput} illustrate the straightforward depth-wise and input-wise dynamic scheme, in which $\mathcal{E}_i$ is implemented as 2D CNN block(s) to support progressive computation. In these cases, We can see naive depth-wise dynamic leaves temporal relation unexplored and input-wise dynamic suffers from limited temporal coverage of input frames when it stops early. 
%In this pamper, we propose joint input-wise and network depth-wise dynamic inference for efficiency improvement. Fig. \ref{fig:dynamic inference} demonstrates several straightforward dynamic inference schemes as well as our proposed framework.  
%As shown in Fig. \ref{fig:addepth}, directly extending the state-of-the-art dynamic depth network called MSDNet \cite{msdnet} to a temporal segment network (TSN) framework \cite{tsn}, the depth-wise dynamic inference can be easily achieved with the early stop trick, but such a framework totally ignores input-wise dynamics of different videos. Another approach to dynamic inference is shown in Fig. \ref{fig:adinput}. It progressively averages classification scores of input frames and breaks the computation flow once the resulted classification score is reliable enough. As we can see, when it stops early, the temporal coverage of input frames is limited due to the sequential input order. This can be harmful to the recognition performance.
To this end, we propose a frame permutation mechanism and leverage online temporal shift module to alleviate these issues.

\subsubsection{Frame Permutation}
%Figure \ref{fig:adinput} shows a straightforward way to implement the input-wise dynamic inference. Based on the naive way, 

Frame permutation scheme can enlarge the temporal coverage of input frames when it stops early and construct diverse temporal strides for input. To exemplify this, suppose eight frames are evenly sampled from a video with a length of $L$, we can see from Fig.\ref{fig:adinput}, if three frames are processed and early stop is made, the temporal coverage of the input $[F_0, F_{1}, F_{2}]$ will be $3L/8$. With frame permutation, the coverage of the input $[F_0, F_{3}, F_{5}]$ changes to $6L/8$ and the temporal strides of $3L/8$ and $2L/8$ are both contained in the inputs, as shown in Fig. \ref{fig:admt-input}.

In this paper, we set $N$ to 8. We start with the case of $L=8$. The extract frames $[F_0, F_1,...,F_7]$ are permuted such that
\begin{equation}
[\mathcal{I}_0,\mathcal{I}_1,...,\mathcal{I}_7] = [F_0, F_3, F_5, F_6, F_1,F_2,F_4,F_7].    
\end{equation}
The temporal order of the shuffled frames no longer holds, and motion noise will be introduced. We divide them into two groups, $[F_0, F_3, F_5, F_6]$ and $ [F_1, F_2, F_4, F_7]$, such that frames in both groups keep in sequential order. 
We can also see there are multiple temporal strides, namely $3L/N, 2L/N, L/N$, in both groups. If temporal relation is modeled inside each group, multiple strides can help to capture better temporal dynamics of different actions whose motion intensities are varying. 
When the number of sampled frames $l$ equals $N\times E$ where $E$ is the size of each frames set, then our permutation mechanism designed for eight frames can be extended as follows. Firstly, the sampled frames are divided into eight sets
\begin{equation}
    \mathcal{F}_i = \{F_{iE},F_{iE+1}, ..., F_{(i+1)E-1}\}, i=0,1,...,7.
\end{equation}
Then, we shuffle the input frames at frame-set granularity
\begin{equation}
[\mathcal{I}_0,\mathcal{I}_1,...,\mathcal{I}_N] = [\mathcal{F}_0, \mathcal{F}_3,\mathcal{F}_5, \mathcal{F}_6, \mathcal{F}_1,\mathcal{F}_2,\mathcal{F}_4,\mathcal{F}_7].    
\end{equation}
The shuffled frame sets are also divided into two groups. Correspondingly, each block of the computation graph is extended to $E$ copies for processing $E$ frames in a set $\mathcal{F}_i$.

%In Fig. \ref{fig:admt-input}, we showcase a permutation strategy when eight input frames are extracted from the video for recognition. If the eight frames are divided into two groups, we can figure out that the permuted input sequence contains multiple temporal strides inside each group, which can be a side-product for better temporal modeling. Our final model (Fig. \ref{fig:unify}) combines input-wise and depth-wise adaption into a whole. Specifically, 

%The input axis and the depth axis composes a 2-dimensional grid, and we predefine $K$ checkpoints on the grid, and the prediction process can be early broken adaptively to enable joint input-wise and depth-wise dynamic testing.

\subsubsection{Online Temporal Shift}
To tackle the problem of missing temporal relation when only 2D convolutions are applied on input frames, we adopt a temporal shift module for efficient temporal modeling. The original temporal shift module~\cite{tsm} shifts part of the channels from feature maps of each input frame forward and backward to that of its adjacent input frames, which facilitates information exchange among neighbouring frames with zero FLOPs cost and negligible time cost. Under dynamic inference circumstance, the original TSM cannot meet the requirement of progressive computation since the final feature maps have to rely on those of its later input frames.
To integrate TSM into our framework, we tailor original TSM approach by only applying forward shift, which makes it possible for online process and it is called \emph{online temporal shift}. 
Specifically, we add an online temporal shift module between $\mathcal{I}_nB_m$ and $\mathcal{I}_{n+1}B_m$ for arbitrary $n$ and $m$ to fully take advantage of multi-stride input brought by the frame permutation. The overall unified framework is shown in Fig.\ref{fig:unify}. 
It is worthy of noting that, our method is a general design, when the $K$ checkpoints are located in the last row or last column, it is specialized to be input-wise or depth-wise dynamic inference framework.

\begin{figure}[t]
      \begin{center}
      \includegraphics[width=0.47\textwidth]{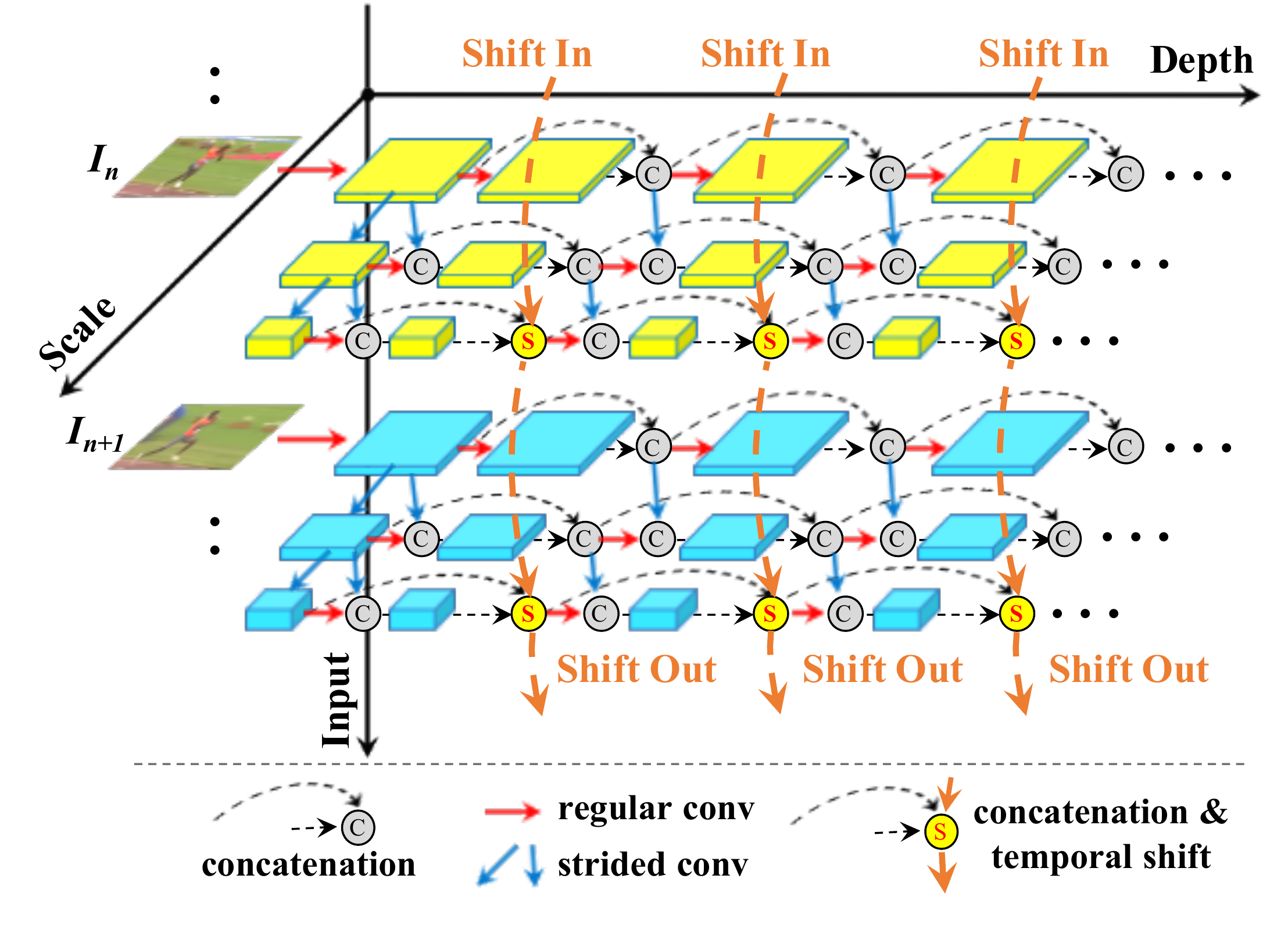}
    %   \vspace{-3 ex}
      \caption{Illustration of the first four layers of our network with three scales. The depth-axis corresponds to the depth of the network, the scale-axis corresponds to the scale of the feature maps, and the input-axis corresponds to the frame of video.}
      \label{fig:layout}
      \end{center}
      \vspace{-3 ex}
\end{figure}

\section{Experiments}
To validate effectiveness of our idea and provide comprehensive understanding on dynamic inference, we implemented three sample frameworks based on state-of-the-art 2D CNN MSDNet-38 \cite{msdnet}, ResNet-50 and ResNet-101 \cite{resnet}, respectively. More deeper and powerful backbones could be leveraged, but here we do not show such experiments because 1) we focus on efficient action recognition and heavy network is not suitable; 2) the three instances are able to achieve proof of concept purpose. MSDNet-38 contains five blocks (M=5) and four scales (S=4) are used in each block. 
Note that the online temporal shift module is only adopted on feature maps at the coarsest scale for all layers except the first layer. Fig. \ref{fig:layout} illustrated the framework using only three scales for convenience. $K$ is 6 for MSDNet and each prediction head is composed of an average pooling and a linear layer. For ResNet, online temporal shift model is added at each residual block and we only append early-exit classifiers to the Res3, Res4 and Res5 and $K$ is set to 4 (the last 4 checkpoints in Fig. \ref{fig:unify}). 
% Each classifier has two down-sampling convolution layers with $3 \times 3$ filters, followed by a average pooling and a linear layer. 
$T_k$ is determined as follows: the \#FLOPs at the $k^{th}$ checkpoint is a known prior $G_k$ and $D$ is the total number of videos. We denote by $0 < q \le 1$ a fixed exit probability that a sample that reaches a checkpoint will obtain a classification with sufficient confidence to exit. We assume that $q$ is constant across all checkpoints, which allows us to compute the probability that a sample exits at the $k^{th}$ checkpoint as: $q_k = {z(1-q)}^{k-1}q$, where $z$ is a normalizing constant that ensures that $\sum_k p(q_k)=1$. Given average inference budget $Q$ in FLOPs, this gives rise to the constraint $\sum_k q_k G_k \le Q$. We can solve this
constraint for $q$ and determine the thresholds $T_k$ on a validation set in such a way that approximately $D* q_k$ validation samples exit at the $k^{th}$ checkpoint.

\subsection{Datasets and Evaluation Metrics}
To comprehensively evaluate our proposed method, we perform extensive experiments on the recent large scale dataset named Kinetics-400~\cite{kay2017kinetics}. We also conduct experiments on heavy temporal relation sensitive datasets, including Something-Something v1 \& v2 \cite{sth-sth}. For these datasets, the actions therein mainly include object-object and human-object interactions, which require strong temporal relation to well categorizing them. Moreover, transfer learning experiments on the UCF-101\cite{ucf101} and HMDB-51\cite{hmdb}, which are much smaller than Kinetics-400, is carried out to show the transfer capability of our solution. %The statistics of these datasets are listed in Table~\ref{t:datasets}.
%For UCF101 and HMDB51, our experiments follow the original evaluation scheme using three training/testing splits and report average class accuracy over these splits.
The evaluation metric is top-1 precision for Kinetics-400 and Something-Something. We also report average FLOPs/Video in the testing phase as well as the number of model parameters to depict model complexity. In this paper, we only use the RGB frames of these datasets for experiments.

%% \vspace{-8mm}
%\begin{table}
%\label{t:datasets}
%\centering
%\begin{tabular}{c|c|c}
%\hline\hline     
%Dataset & Classes & Videos  \\ 
%\hline  
%Kinetics400 & 400 & 306,245 \\
%Something-V1 / V2 & 174 & 108,499 / 220,847  \\ 
%UCF101 & 101 & 13,320  \\
%HMDB51 & 51 & 6,766 \\
%\hline\hline
%\end{tabular}
%\caption{Statistics of the datasets used in our experiments.}
%% \vspace{-10mm}
%\end{table}

\subsection{Implementation Details}

\textbf{Training} Data augmentation and preprocessing strategy is the same as TSN \cite{tsn}. In our experiments, $l$ is set to 16. 
To evaluate the proposed network on several action recognition datasets, we pre-trained MSDNet-38, ResNet-50, ResNet-101 on the ImageNet-1k \cite{deng2009imagenet} for initialization. The loss is the sum of cross-entropy at each checkpoint. For Kinetics-400, Something-Something v1 \& v2, we start training with a learning rate of 0.01 and reduce it by a factor of 10 at 25, 35, 45 epochs and stop at 50 epochs. Dropout ratio is 0.5. Since UCF-101 and HMDB-51 are not large enough and are prone to over-fitting, we followed the common practice to use Kinetics pre-trained model as initialization and in total 25 epochs are trained, with an initial learning rate of 0.01 and it is decayed by 10x every ten epochs. Higher dropout ratio of 0.8 is used. Stochastic gradient descent (SGD) with a mini-batch size of 128 is utilized as an optimizer, and its momentum and weight decay value is set to 0.9 and 5e-4, respectively. BatchNorm layers \cite{bn} are all finetuned during training.

\textbf{Inference} %Many state-of-the-art methods rely on some sophisticated testing strategy or post-processing techniques, such as 10-crop testing, to boost performance. 
We apply single centre 224x224 cropping to predict the class labels and the 16 frames are evenly sampled from videos. 
%Under the dynamic inference circumstance, the prediction will early stop at some checkpoint once the classification score of its corresponding prediction head is no less than a pre-defined threshold of $T$. Conditioned on our assumption that ``easier'' videos require fewer input frames and (or) shallower network depth, when $T$ gets smaller, more videos will run into the earlier exit and otherwise fewer videos. %If the average inference FLOPs budget for a video is given, we can find an appropriate threshold of $T$ to satisfy the computation cost limitation and get the optimal recognition performance.  
To get the accuracy-FLOPs curves in the paper, we traverse different budget $Q$ to determine $T_k$ on the train set and measure the actual accuracy and FLOPs. %we can obtain the distribution of number of videos which stop at each checkpoints. The \#FLOPs at each checkpoint is a known prior, so we can calculate the average \#FLOPs and the corresponding top-1 accuracy. 
As for validation set, we use the critical point of $Q^*$, where accuracy drops little and FLOPs are smaller than full inference, obtained from the accuracy-FLOPs curve on the validation set to determine $T_k^*$ for testing and then accuracy and average \#FLOPs per video is calculated.

\subsection{Empirical Analysis} 
\begin{figure}[!t]
\begin{center}
\subfigure[Kinetics-400]{
\includegraphics[width=0.47\columnwidth]{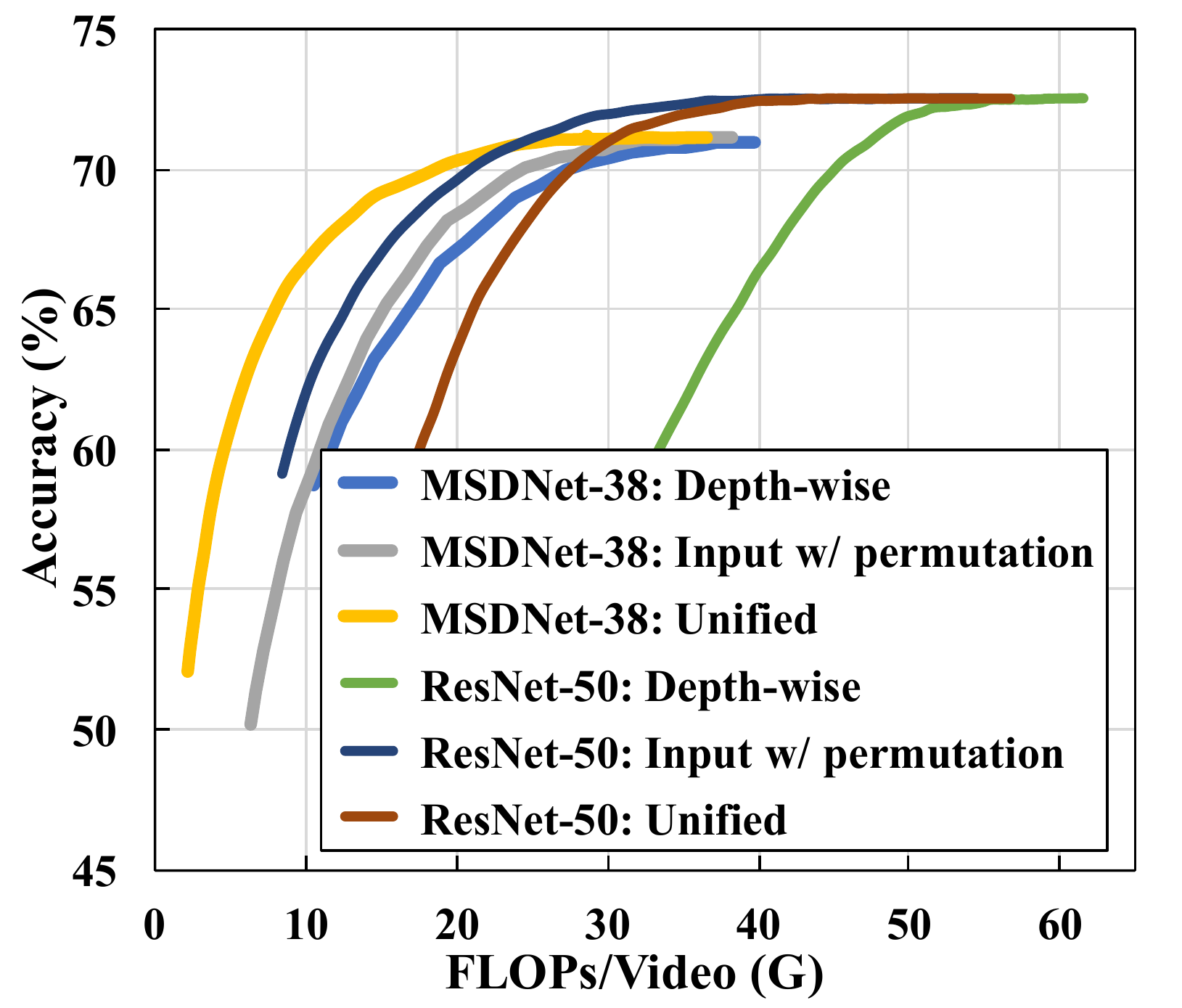}
\label{fig:k400-pk}
}
\subfigure[Something-Something v2]{
\includegraphics[width=0.47\columnwidth]{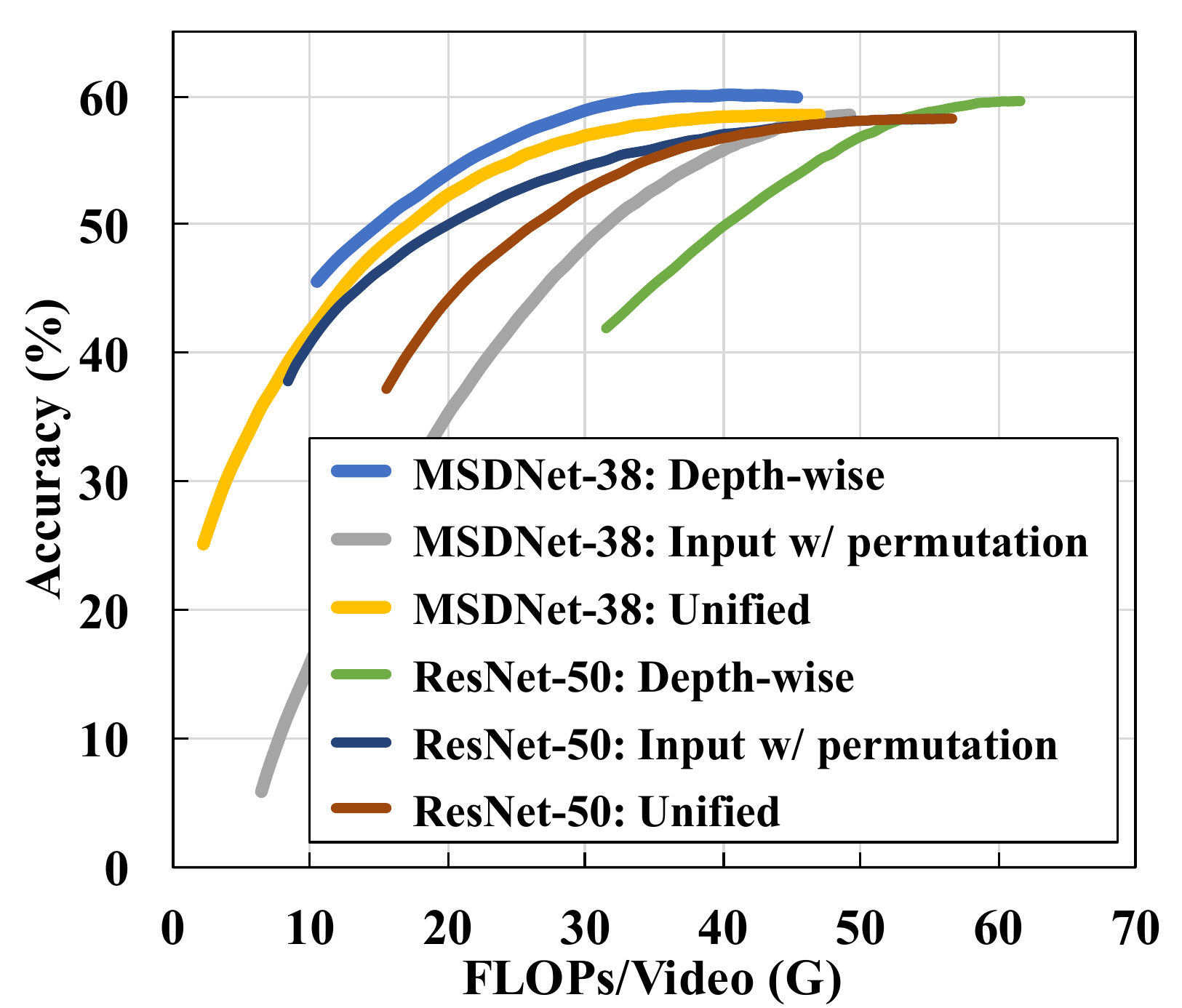}
\label{fig:sthv2-pk}
}
\end{center}
\vspace{-2 ex}
\caption{MSDNet-38 v.s. ResNet-50 on two types of datasets.}
\label{fig:backbone-pk}
\end{figure}
\textbf{MSDNet v.s. ResNet} First of all, we would like to know how backbone makes a difference. Both Kinetics-400 and Something-Something v2 is used to evaluate the dynamic inference mechanisms proposed in Section~\ref{sec:approach} using the RGB modality. As shown in Fig. \ref{fig:k400-pk}, MSDNet-38 performs better than ResNet-50 in depth-wise dynamic inference owing to the multi-scale feature maps and dense connectivity. % as well as the larger spatial size of features at shallow layers of ResNet-50. 
Influenced by the limited performance of depth-wise dynaimc inference, joint depth-wise and network depth-wise solution performed worse than input-wise dynamic inference for ResNet-50. %Nevertheless, the input-wise dynamic inference performs well for ResNet-50 on Kinetics-400. 
On Something-Something v2, similar observations can be found.

\begin{figure}[ht]
\begin{center}
\subfigure[W/o online temporal shift]{
\includegraphics[width=0.47\linewidth]{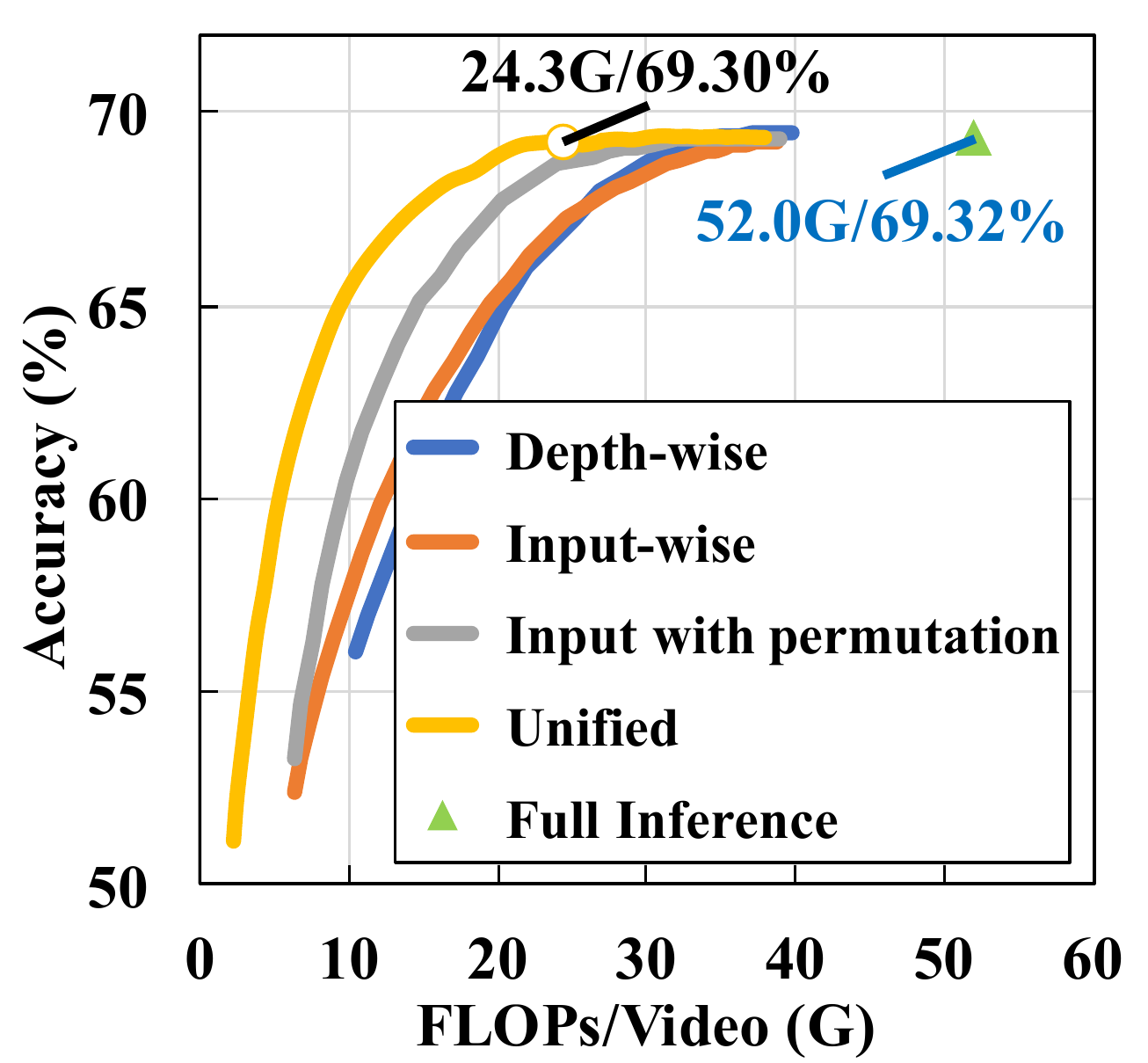}
\label{fig:K400-TSN}
}
\subfigure[With online temporal shift]{
\includegraphics[width=0.47\linewidth]{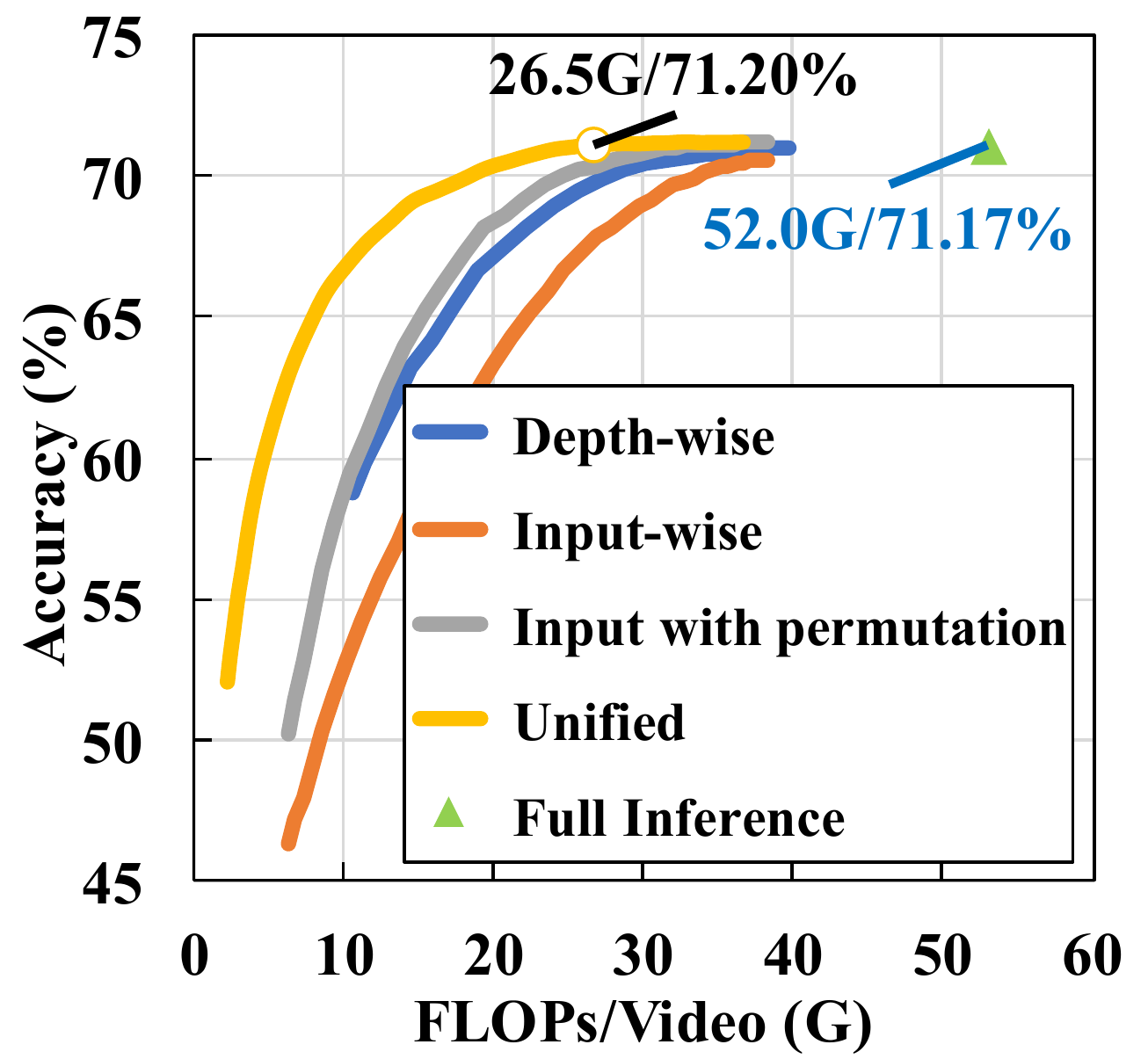}
\label{fig:K400-TSM}
}
\end{center}
\vspace{-2 ex}
\caption{Ablation study  with MSDNet backbone on Kinetics-400 dataset. ``Full Inference'' means that, for each video, only the prediction head of the last checkpoint is used.}
\label{fig:K400}
\end{figure}

\begin{figure}[ht]
\begin{center}
\subfigure[W/o online temporal shift]{
\includegraphics[width=0.47\columnwidth]{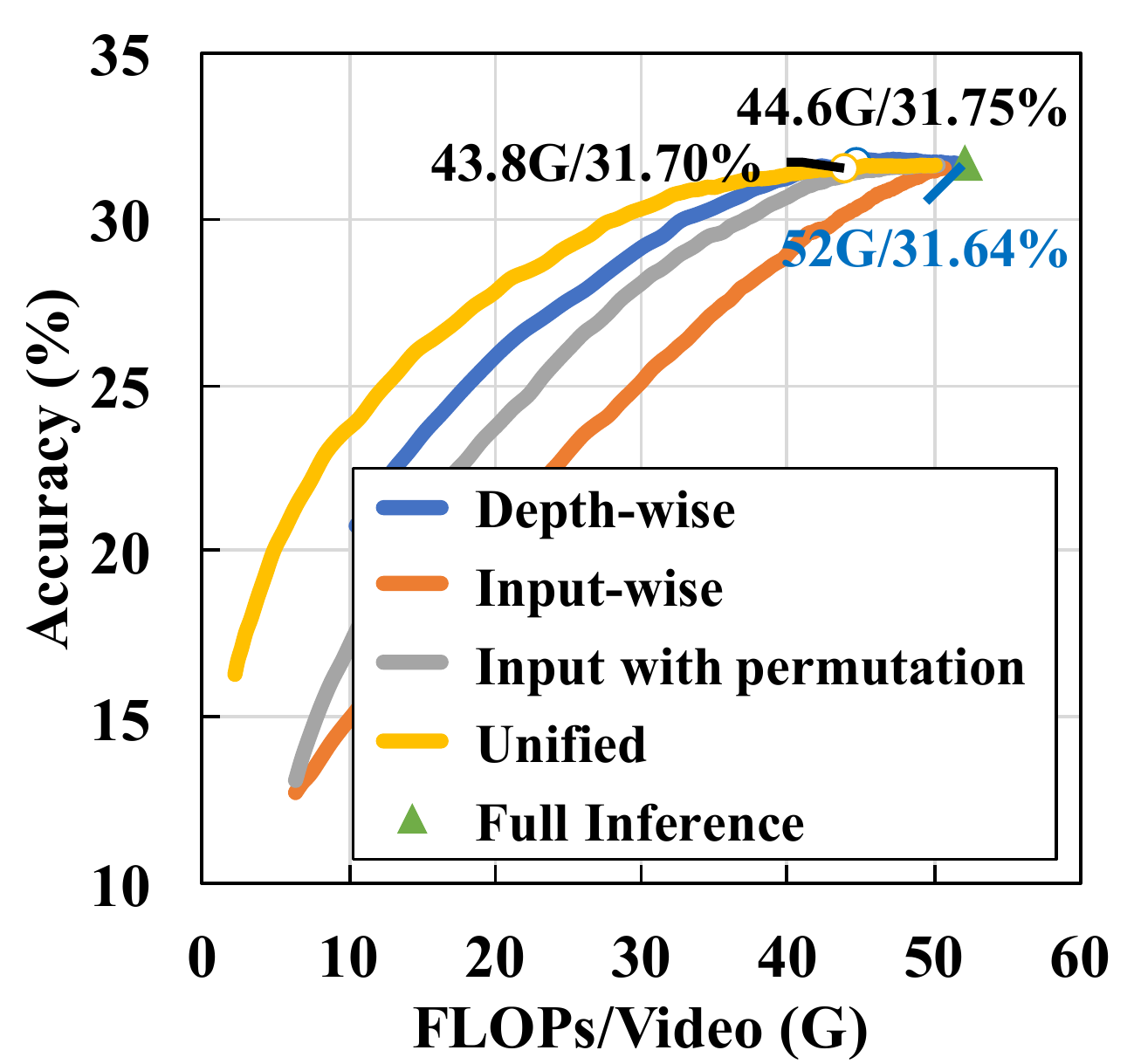}
\label{fig:sthv2-TSN}
}
\subfigure[With online temporal shift]{
\includegraphics[width=0.47\columnwidth]{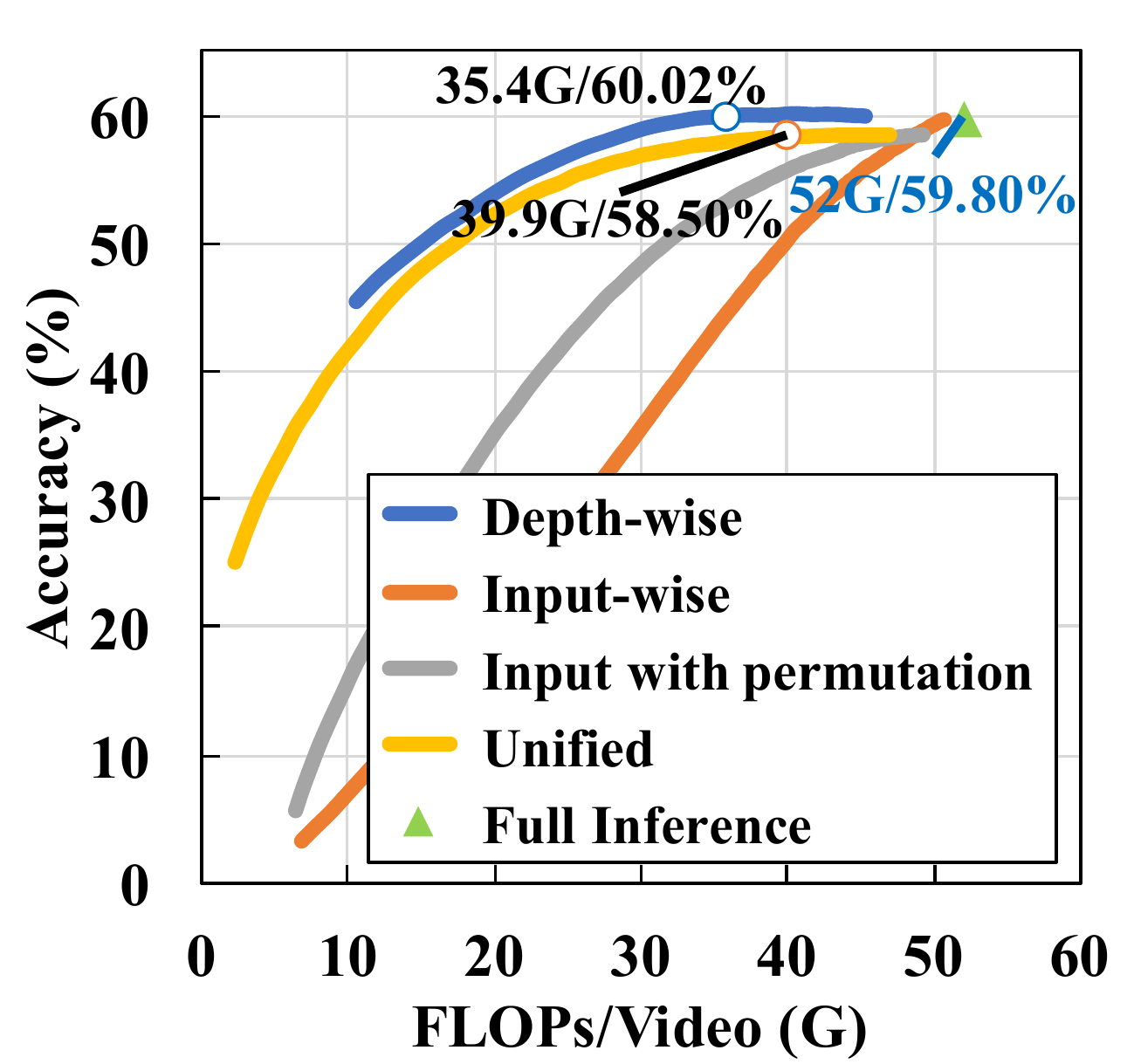}
\label{fig:sthv2-TSM}
}
\end{center}
\vspace{-2 ex}
\caption{Ablation study with MSDNet backbone on Something-Something v2 dataset.}
% \vspace{-2 ex}
\label{fig:sth}
\end{figure}

\textbf{Ablation studies} Fig. \ref{fig:K400} and Fig. \ref{fig:sth} show the results based on MSDNet under different dynamic inference settings. Accordingly, the effectiveness of each component in our framework is analyzed as follows.

\textbf{Depth-wise dynamic inference} As shown in Fig.~\ref{fig:K400} and Fig.~\ref{fig:sth}, it can be concluded from the results that depth-wise dynamic inference is effective on both datasets no matter whether the online temporal shift is enabled or not. It is a very interesting observation from the results of Fig. \ref{fig:sthv2-TSM} that depth-wise dynamic inference with online temporal shift on Something-Something v2 performs even slightly better the joint input-wise and depth-wise adaptation. This is reasonable because temporal relation is crucial for this dataset, input adaptation brings very little gain such that network trained with pure depth-wise adaptation can be more competitive than its joint depth-wise and input-wise version. We can see that on Something-Something v2, without and with the online temporal shift, depth-wise adaptation achieves top-1 accuracy of 31.75\% and 60.02\%, and the \#FLOPs can be reduced to 44.6G and 35.4G, respectively.

\textbf{Input-wise dynamic inference and frame permutation} We find that on Something-Something v2, input-wise adaptation brings negligible FLOPs saving, while on Kinetics-400 it can save FLOPs to some extent.  
In Fig.~\ref{fig:K400} and Fig.~\ref{fig:sth}, on both datasets, it is clear that frame permutation improves input-wise adaptation, regardless of whether online temporal shift module is used or not. Even though temporal relation is not so crucial on Kinetics-400 compared to Something-Something v2, frame permutation still gains a lot, and we can infer that temporal coverage of the input sequence is essential to video classification. 
Naturally, other architectures such as TSM, can also benefit from the frame permutation mechanism for dynamic inference.

\textbf{Online temporal shift} Temporal modeling by online temporal shift is also essential for the final performance. It is seen from the figures, the online temporal shift improves the top-1 accuracy upper bound of our framework from 69.32\% to 71.17\% and 31.64\% to 59.80\% on Kinetics-400 in Fig.~\ref{fig:K400} and Something-Something v2 in Fig.~\ref{fig:sth}, respectively.

\textbf{Joint input-wise and depth-wise dynamic inference} On Kinetics, the proposed joint solution benefits from an adaption of both dimensions in Fig.~\ref{fig:K400}. The \#FLOPs is reduced to 24.3G and 26.5G from 52G without or with the online temporal shift, respectively. We also show the distribution of videos predicted at each checkpoint in Table.\ref{t:n_video}. However, in Fig.~\ref{fig:sth}, on Something-Something v2, it performs comparably with or even slightly worse than depth-wise adaption owing to the heavy temporal dependency characteristic of the dataset and input-wise adaptation can hardly bring any gain.

%%%%%%%%%%%%%%%%%%%%%%For K400%%%%%%%%%%%%%%%%%%%%%%
\begin{table*}[t]
\begin{center}
\begin{adjustbox}{max width=\textwidth}
\begin{threeparttable}
\begin{tabular}{c|c|c|c|c|c}
\hline\hline
Framework & Backbone & Input $\times$ \# Clips & Prec@1 & \# Params & FLOPs/Video \tabularnewline
\hline\hline
I3D~\cite{i3d} & 3D BN-Inception & [All$\times$3$\times$256$\times$256]$\times$1  &71.1 &12.7M &544.44G \tabularnewline
\cline{1-3}\cline{4-6}

%\multirow{2}{*}{I2D \cite{nonlocal}} & \multirow{2}{*}{ResNet-50} & [32$\times$3$\times$256$\times$256]$\times$1  & 62.42 & \multirow{2}{*}{24.27M} & 26.29G  \tabularnewline
%&  & [32$\times$3$\times$256$\times$256]$\times$10  &$69.90$ & &262.9G  \tabularnewline
  %\cline{1-3}\cline{5-7}
  %     TSN             & InceptionResnet-V2 & [25$\times$3$\times$331$\times$331]$\times$1 &         &73.01 & &  \\
%\cline{1-3}\cline{4-6}

%\multirow{2}{*}{I3D \cite{nonlocal}} & \multirow{2}{*}{3D ResNet-50} & [32$\times$3$\times$256$\times$256]$\times$1 &64.65 & \multirow{2}{*}{35M} &164.84G \tabularnewline
%& & [32$\times$3$\times$256$\times$256]$\times$10   &71.86 & &1648.4G \tabularnewline
%\cline{1-3}\cline{4-6}

S3D \cite{s3d} & 3D BN-Inception & [All$\times$3$\times$224$\times$224]$\times$1 &72.20 &8.8M &518.6G \tabularnewline
\cline{1-3}\cline{4-6}

ARTNet with TSN \cite{ARN} & 3D ResNet-18 & [16$\times$3$\times$112$\times$112]$\times$250   &69.2 &35.2M &5925G \tabularnewline
\cline{1-3}\cline{4-6}

\multirow{2}{*}{MF-Net \cite{mfnet}} & \multirow{2}{*}{-} & [16$\times$3$\times$224$\times$224]$\times$1 &65.00 & \multirow{2}{*}{8.0M} &11.1G \tabularnewline
& & [16$\times$3$\times$224$\times$224]$\times$50 &72.80 &  &555G \tabularnewline
\cline{1-3}\cline{4-6}

ECO \cite{eco} & BN-Inception+3D ResNet-18 & [16$\times$3$\times$224$\times$224]$\times$1  &69.00 &47.5M &64G \tabularnewline
\cline{1-3}\cline{4-6}

R(2+1)D RGB \cite{r2+1d} & ResNet-34 & [32$\times$3$\times$112$\times$112]$\times$10 &72.00 &63.8M &1524G \tabularnewline
\cline{1-3}\cline{4-6}

\multirow{2}{*}{Nonlocal-I3d~\cite{nonlocal}} & \multirow{2}{*}{ResNet-50} & [128$\times$3$\times$224$\times$224]$\times$1 &67.30 & \multirow{2}{*}{35.33M} &145.7G \tabularnewline
& & [128$\times$3$\times$224$\times$224]$\times$30 & 76.50 & &4371G \tabularnewline
\cline{1-3}\cline{4-6}
\hline\hline

\multirow{3}{*}{TSN RGB \cite{tsn}} & BN-Inception & [25$\times$3$\times$112$\times$112]$\times$10  &69.1 & 10.7M &500G \tabularnewline
& ResNet-50 & [8$\times$3$\times$224$\times$224]$\times$1 & 66.80 & 24.3M &33G \tabularnewline
& ResNet-50 & [16$\times$3$\times$224$\times$224]$\times$1 & 67.80 & 24.3M &64G \tabularnewline
\cline{1-3}\cline{4-6}

\multirow{2}{*}{TSM \cite{lin2018tsm}} 
& \multirow{2}{*}{ResNet-50} & [8$\times$3$\times$224$\times$224]$\times$1 & 70.60 & 24.3M &33G \tabularnewline
& & [16$\times$3$\times$224$\times$224]$\times$1 & 72.50 & 24.3M &64G \tabularnewline
\cline{1-3}\cline{4-6}

\multirow{2}{*}{StNet~\cite{stnet}} &ResNet-50 & [25$\times$15$\times$256$\times$256]$\times$1  &69.85 &33.16M &189.29G \tabularnewline
& ResNet-101 & [25$\times$15$\times$256$\times$256]$\times$1  &71.38 &52.15M &310.50G \tabularnewline
\cline{1-3}\cline{4-6}

\multirow{4}{*}{\textbf{Proposed}} & MSDNet-38 (Full) & [16$\times$3$\times$224$\times$224]$\times$1  & 71.17 & 62.31M & 52G \tabularnewline
 & MSDNet-38  & [16$\times$3$\times$224$\times$224]$\times$1 & \textbf{71.20} &  62.31M & \textbf{26.5G} \tabularnewline
  & ResNet-50  & [16$\times$3$\times$224$\times$224]$\times$1 & \textbf{72.57} &  29.12M & \textbf{35G} \tabularnewline
  & ResNet-101  & [16$\times$3$\times$224$\times$224]$\times$1 & \textbf{74.70} &  48.12M & \textbf{66G} \tabularnewline
\cline{1-3}\cline{4-6}  

\hline\hline
\end{tabular}
\end{threeparttable}
\end{adjustbox}
\end{center}
\caption{Comparison of our method with several state-of-the-art 2D/3D convolution-based solutions. The results are reported on the validation set of Kinetics-400, with RGB modality only. We investigate both Prec@1 and model efficiency w.r.t. the total number of model parameters and FLOPs needed in inference. Here, ``All'' denotes using all frames in a video. }
\label{t:k400}
\end{table*}
%%%%%%%%%%%%%%%%%%%%%%For K400 end%%%%%%%%%%%%%%%%%%%%%%

\subsection{Comparison with State-of-the-arts}
\textbf{Results on Kinetics-400}
We evaluate the proposed framework against the recent state-of-the-art 2D/3D convolution-based solutions. Extensive experiments are conducted on the Kinetics-400 to compare all models in terms of their effectiveness (i.e., top-1 accuracy) and efficiency (reflected by the total number of model parameters and FLOPs needed in the inference phase). Results are summarized in Table \ref{t:k400}. To be noted, the latest camera ready version of TSM reported results of these datasets using dense sampling for higher performance. Hence, in this paper we cited the results of these datasets using uniform sampling reported in its arXiv v1 version~\cite{lin2018tsm}.

The results show that our proposed dynamic inference with MSDNet-38 backbone achieves top-1 precision of 71.20\% with single RGB modality, and it requires only 26.5G FLOPs per video on average. When large backbone models are used, our method still strike good performance-FLOPs trade-off. Compared to StNet-Res101, our method with ResNet-50 exhibits better recognition performance (72.57\% v.s. 71.38\%), but the \#FLOPs are significantly reduced by over $10\times$ (from 310G to 35G). When compared to the recent state-of-the-art ECO and TSM-Res50, our model improves the accuracy from 69\% and 72.5\% to 72.57\%, meanwhile the \#FLOPs is reduced from 64G and 64G to 35G, respectively. Other methods, such as MF-Net, S3D, and R(2+1)D, though their marginal performance gains are obtained at the price of at least 10 times of computation cost, our method with ResNet-101 still achieve better recognition performance than them while \#FLOPs is only 66G. In brief, our model achieves excellent performance-cost trade-off with different backbones.

\begin{table}%[ht]
\begin{center}
\begin{adjustbox}{max width=\linewidth}
% \begin{threeparttable}
\begin{tabular}{ c | c | c | c | c | c| c| c}
\hline\hline
Checkpoint & 1 & 2 & 3 & 4 & 5 & 6 & Total\\
\hline
\#Videos & 1993 & 2392 & 2871 & 3445 & 4134 & 4961 & 19796 \\ 
% ResNet-50 & - & - & 1.89 & 3.36 & 4.10  \\ \hline

% ResNet-50 & - & 0.86 & 1.96 & 3.51 & 4.32  \\
% FLOPs (G) & 0.61 & 1.43 & 2.28 & 2.96 & 3.25  \\
% MSDNet-43 & 0.75G / 1.77G / 2.82G / 3.62G / 4.01G  \\
\hline\hline
\end{tabular}
% \end{threeparttable}
\end{adjustbox}
\end{center}
\caption{Number of videos which stop at each checkpoint on Kinetics-400 validation set. }
\vspace{-2 ex}
\label{t:n_video}
\end{table}

%%%%%%%%%%%%%%%%%%%%%%%sth-sth%%%%%%%%%%%%%%%%%%%%%%
\begin{table*}%[t]
	\begin{center}
	\begin{adjustbox}{max width=\textwidth}
		\begin{threeparttable}
			\begin{tabular}{c|c|c|c|c|c|c|c|c|c}
				\hline\hline
				\multirow{2}{*}{Method} & \multirow{2}{*}{Backbone} & \multirow{2}{*}{Pretrain} & \multirow{2}{*}{FLOPs/Video} & \multicolumn{2}{c|}{Something-Something v1} & \multicolumn{4}{c}{Something-Something v2}\tabularnewline
				\cline{5-10}
				& &  &  &  top-1 val & top-5 val & top-1 val & top-5 val & top-1 test & top-5 test\tabularnewline
				\hline
				\hline
				ECO$_{16F}$  & BNInception+ & \multirow{2}{*}{Kinetics} & 64G & 41.4 & -  & - & - & - & -\tabularnewline
				
				ECO$_{EN}Lite$  & 3D ResNet-18 & & 267G & 46.4 & -  & - & - & - & -\tabularnewline

				\hline
				I3D  & \multirow{2}{*}{3D ResNet-50}  &  \multirow{2}{*}{Kinetics} & 306G & 41.6 & 72.2 & -  & - & - & -\tabularnewline
				Non-local I3D+GCN  &  &   & 606G & 46.1 & 76.8 & - & - & - & -\tabularnewline
				\hline
				\hline
				TSN$_{8F}$ & \multirow{2}{*}{ResNet-50} \multirow{2}{*}{} & \multirow{2}{*}{Kinetics} & 33G & 19.7 & 46.6 & 27.8 & 57.6 & - & -\tabularnewline
			   	TSN$_{16F}$	& &  & 65G & 19.9 & 47.3  & 30.0 & 60.5 & - & -\tabularnewline
				\hline
				TRN Multiscale  & \multirow{2}{*}{BNInception} &   \multirow{2}{*}{ImageNet} & 33G & 34.4 & - & 48.8 & 77.6 & 50.9 & 79.3\tabularnewline
				TRN Two-Stream  &  & & - & 42.0 & - & 55.5 & 83.1 & 56.2 & 83.2\tabularnewline
				\hline
				TSM$_{8F}$ & \multirow{2}{*}{ResNet-50} \multirow{2}{*}{} & \multirow{2}{*}{Kinetics} & 33G & 43.4 & 73.2 & 58.2 & 84.8 & - & -\tabularnewline
			   	TSM$_{16F}$	& &  & 65G & 44.8 & 74.5 & 58.7 & 84.8 & 59.9 & 85.9\tabularnewline
		        \hline
				\multirow{2}{*}{\textbf{Proposed}} & ResNet-50 &  \multirow{2}{*}{ImageNet} & \textbf{52.8(v1)/48.0G(v2)} & \textbf{45.2}  & \textbf{75.2} & \textbf{58.2} & \textbf{85.2} & - & - \tabularnewline
				& MSDNet-38 & & \textbf{38.4G(v1)/35.4G(v2)} & \textbf{46.5}  & \textbf{75.6} & \textbf{60.0} & \textbf{86.2} & \textbf{60.1} & \textbf{86.6} \tabularnewline
				\hline\hline
			\end{tabular}%
		\end{threeparttable}
	\end{adjustbox}
	\end{center}
	\caption{Performance and FLOPs consumptions of our method on the Something-Something v1 and v2 datasets compared with the state-of-the-art methods. FLOPs/Video are averaged over all the videos from the validation set. }
\label{some}
 \vspace{-2 ex}
\end{table*}

\textbf{Results on Something-Something v1 \& v2}
The Something-Something datasets are more complicated than Kinetics, the comparison of our solution against existing state-of-the-arts are list in Table~\ref{some}. It is observed that, with 16 frames as input, our solution achieves state-of-the-art top-1 accuracy of 46.5\% on Something-Something v1 validation set and 60.02\% on v2. The \#FLOPs of our model is 38.4G/35.4G on v1 and v2 respectively, which is much smaller than ECO$_{EN}lite$, Non-local I3D+GCN and TSM$_{16F}$. Compared to TSM on both datasets, our method achieves better recognition performance than TSM$_{16F}$ while the \#FLOPs is comparable with TSM$_{8F}$. We have also submitted the testing results of Something-Something v2 and results show that the testing performance of our solution is as high as $60.16\%$ in terms of top-1 accuracy.

\subsection{Transfer Learning on UCF-101 \& HMDB-51}
\begin{figure}
\begin{center}
\subfigure[UCF101 Split1]{
\includegraphics[width=0.47\linewidth]{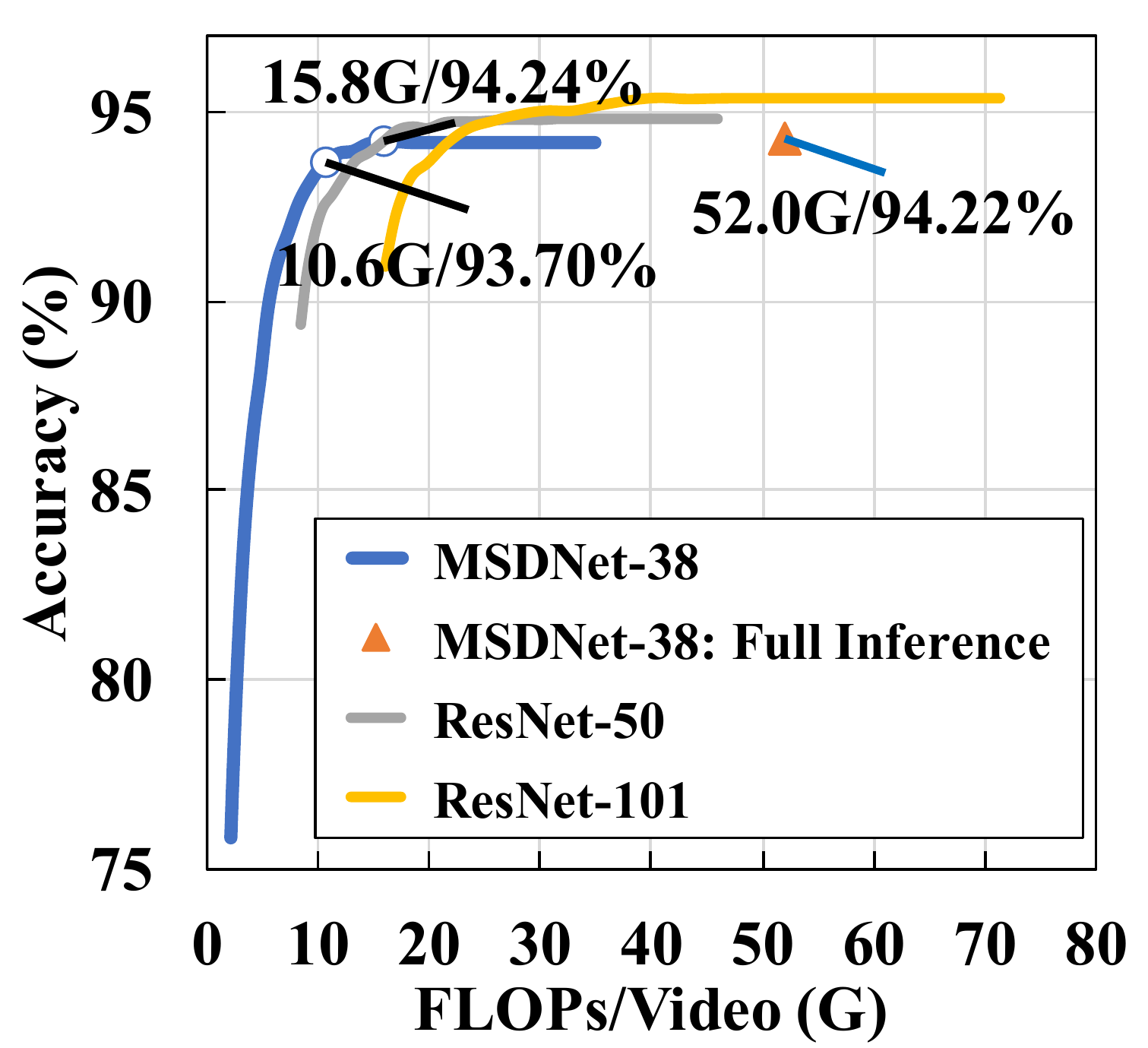}
\label{fig:ucf-s1}
}
\subfigure[HMDB51 Split1]{
\includegraphics[width=0.47\linewidth]{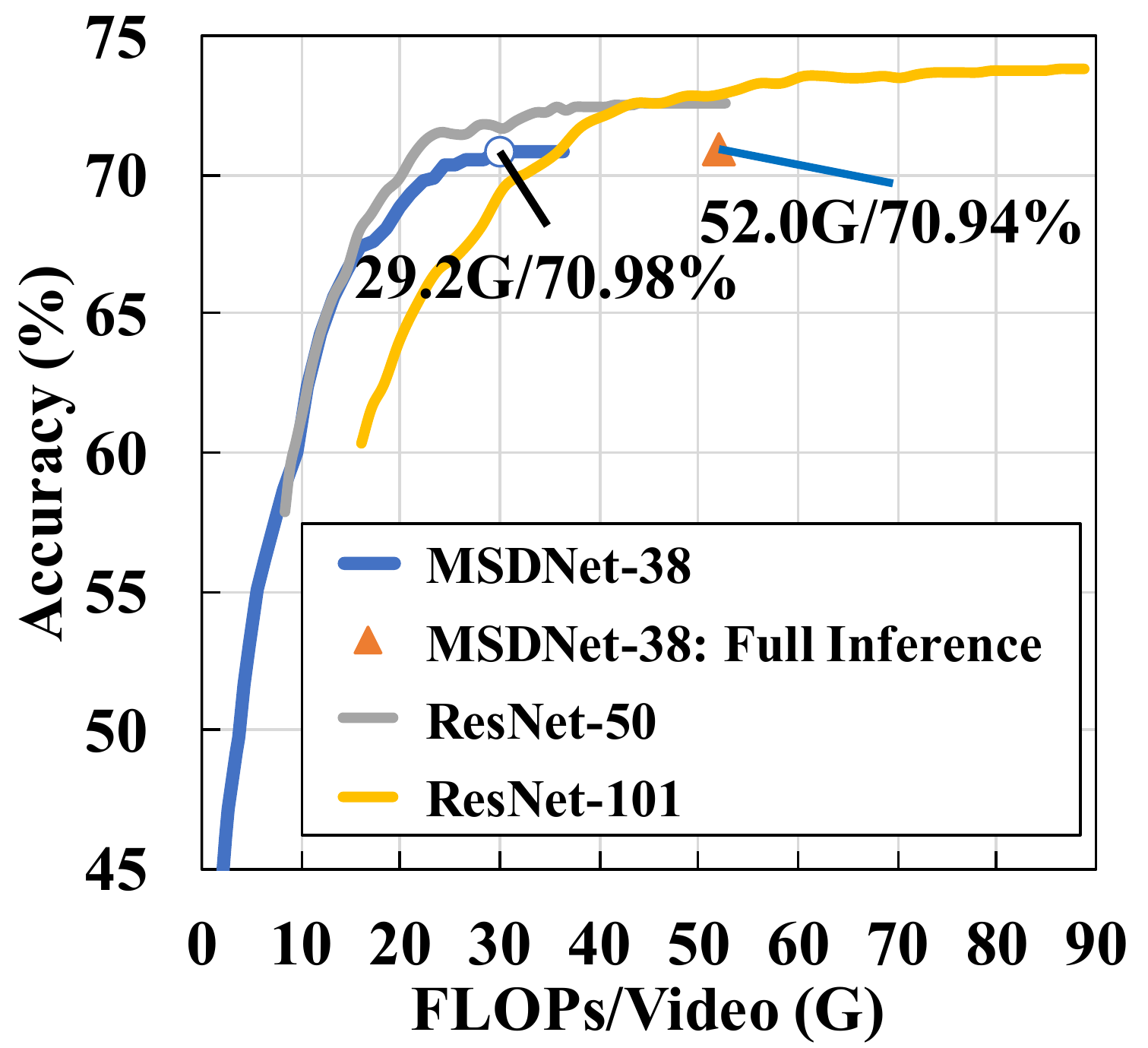}
\label{fig:hmdb-s1}
}
\end{center}
\vspace{-2 ex}
\caption{The Accuracy-FLOPs curves achieved on the first split of UCF101 and HMDB51. RGB modality is used and the models are pretrained on Kinetics-400.}
\label{fig:UCF-HMDB}
\vspace{-2 ex}
\end{figure}

%%%%%%%%%%%%%%%%%%%%%%For ucf101&hmdb51 %%%%%%%%%%%%%%%%%%%%%%
\begin{table}
\begin{center}
\begin{adjustbox}{max width=\linewidth}
\begin{threeparttable}
\begin{tabular}{c|c|c|c|c}
\hline\hline
Method  & Backbone  & FLOPs & UCF-101  & HMDB-51\tabularnewline
\hline\hline
ARTNet with TSN  & 3D ResNet-18   & 5925G & 94.3  & 70.9\tabularnewline
\hline
\multirow{2}{*}{ECO}   & BNInception+    & \multirow{2}{*}{64G} & \multirow{2}{*}{92.8}  & \multirow{2}{*}{68.5}\tabularnewline
&  3D ResNet-18    &  &   & \tabularnewline
\hline
I3D RGB   & 3D Inception-v1   & 544G & 95.1  & 74.3\tabularnewline
\hline
\hline
%TSN   & ResNet-50    & 500G & 86.2  & 54.7\tabularnewline
%\hline
TSN RGB  & BNInception  & 500G & 91.1  & -\tabularnewline
TSN$_{8F}$  & \multirow{2}{*}{ResNet-50} &  33G & 91.5 & 63.2\tabularnewline
TSN$_{16F}$ &   & 64G & 91.4  & 63.6\tabularnewline
\hline
TSM$_{8F}$  & \multirow{2}{*}{ResNet-50} & 33G & 94.0 & 70.3\tabularnewline
TSM$_{16F}$ &  & 64G & 94.5  & 70.7\tabularnewline
\hline
StNet  & ResNet-50 & 53G & 93.5  & -\tabularnewline
\hline
\multirow{6}{*}{\textbf{Proposed}} & \multirow{2}{*}{MSDNet-38} & \textbf{15.8G}  & \textbf{94.2}  & - \tabularnewline
     & & \textbf{29.2G} & - & \textbf{70.1} \tabularnewline
 & \multirow{2}{*}{ResNet-50} & \textbf{18.5G}  & \textbf{94.7}  & - \tabularnewline
     & & \textbf{34.4G} & - & \textbf{72.34} \tabularnewline
 & \multirow{2}{*}{ResNet-101} & \textbf{34.6G}  & \textbf{95.3}  & - \tabularnewline
     & & \textbf{69.1G} & - & \textbf{73.48} \tabularnewline     
     
\hline\hline
\end{tabular}
\end{threeparttable}
\end{adjustbox}
\end{center}
\caption{Transfer leanring performances with RGB modality. %RGB frames are used for training and testing.
Note that the \#FLOPs herein takes into account the testing strategies, such as multi-crop testing, for different models.}
\label{t:UCFHMDB}
\end{table}

We transfer the proposed models pre-trained on Kinetics-400 to the much smaller datasets of UCF-101 and HMDB-51 to show that our method can be well generalized to other datasets. Fig. \ref{fig:UCF-HMDB} show the Mean Class Accuracy v.s. Average FLOPs/Video curve on the split1 of UCF-101 and HMDB-51, respectively. For UCF-101 split1, the average FLOPs can be reduced by 70\% while remaining the same performance with full inference. It is also be observed with 1\% accuracy degradation, around 80\% reduction in average FLOPs is achieved. For HMDB-51, our dynamic inference scheme keep classification accuracy while reducing up to 45\% average FLOPs. Besides, the accuracy on UCF-101 is much higher than HMDB-51, and we believe that our method can reduce more average FLOPs on easier dataset. We also observe that the curve of HMDB-51 is similar to that of Kinetics-400 because the accuracy of our model is very close on these two scene-based datasets.

When compared with other methods, our model is evaluated by following the common practice of averaging accuracy/FLOPs over three training/testing splits of both datasets. The evaluation results can be found in Table~\ref{t:UCFHMDB}. From this table, we can see that our model shows a pretty transfer capability. It obtains pretty good accuracy on UCF-101 and HMDB-51, and the mean class accuracy is 94.2\% and 70.1\%, respectively, which is comparable to or even better than performances of many state-of-the-art solutions. Besides, our model achieves great efficiency improvement, and the average FLOPs/Video can be significantly reduced to 15.8G and 29.2G on UCF-101 and HMDB-51. The results of ResNet based models can be analyzed similarly.

\section{Discussion}
\textbf{Parallelism} Dynamic inference, compared with fixed inference, could degrade the parallelism capability to some extent, but it can be paralleled. The depth-wise dynamic inference can be paralleled in the input dimension, meanwhile input-wise dynamic inference can be paralleled in the input dimension when the computation process is going from one checkpoint to the next one and more than one frames are fed. As for the joint input-wise and depth-wise scheme, it's parallelism capability is further limited, take Fig.\ref{fig:unify} for illustration, if the computation process goes from the $4^{th}$ checkpoint to the $5^{th}$ one, the last block outputs of $[F_0, F_3, F_5, F_6]$ can be calculated in parallel. When it travels from the $5^{th}$ checkpoint to the last one, the whole feature extraction process of $[F_1, F_2, F_4, F_7]$ can be paralleled. 
%We have to admit that the joint input-wise and depth-wise scheme discourages parallelism to some extent, which we leave as one of future research directions. 

\textbf{Optimality} In this paper, our key insight is dynamic inference and we formally summarize its general idea and provide sample instantiating strategies to prove its feasibility. The three framework instances are designed to thoroughly analyze effectiveness of our idea. However, the optimality of these frameworks are not necessarily guaranteed. For example, how these thresholds $T_k$ can be adaptively determined for different videos at different checkpoints; how the checkpoints can be dynamically located from video to video and so on. We believe that all of these problems deserve future research effort. %how the loss function can be designed to take into consideration that the feature of video predicted by different checkpoints exhibits different distribution. 
%However, there are still many open questions unresolved in this work, which are left as our future work. For instance,  how the checkpoints can be dynamically located from video to video and how the loss function can be designed to take into consideration that the feature of video predicted by different checkpoints exhibits different distribution. 

\section{Conclusion}
In this paper, we focus on improving action recognition efficiency in videos by dynamic inference, which takes advantage of distinguishability variation among different videos. Both the proposed frame permutation and online temporal shift scheme contribute largely to performance improvement. By exploiting dynamic inference, our models are verified on multiple well-known datasets being able to significantly save average FLOPs consumption per video while remaining recognition performance unaffected. To our best knowledge, we are the first to propose dynamic inference for efficient video recognition, and it is proven to be a promising direction.

{\small
\bibliographystyle{ieee_fullname}
\bibliography{egbib}
}

\end{document}